\def\year{2017}\relax
\documentclass[letterpaper]{article}
\usepackage{aaai17}
\usepackage{times}
\usepackage{helvet}
\usepackage{courier}
\usepackage{dcolumn}
\usepackage{graphics}
\usepackage{helvet}
\usepackage{courier}
\usepackage{amssymb}
\usepackage{amsmath}
\usepackage{array}
\usepackage{float}
\usepackage{graphicx,epsfig,epstopdf}
\usepackage{psfrag}
\usepackage{subfigure}
\usepackage{tikz,rotating}
\usepackage{url}
\usepackage{color}
\usepackage{theorem}
\usepackage{algorithm}
\usepackage[noend]{algpseudocode}
\usepackage{verbatim}
\usepackage{enumitem}
\usepackage{multirow}
\usepackage{mathabx}
\usepackage{xr}
\usepackage{appendix}

\newtheorem{theorem}{Theorem}

\newtheorem{definition}{Definition}

\begin{document}

\title{Team--maxmin equilibrium: efficiency bounds and algorithms}
\author{Nicola Basilico\\ ~ University of Milan\\ ~ Via Comelico, 39/41
\\ ~ Milano, Italy\\ ~ nicola.basilico@unimi.it \And Andrea Celli, Giuseppe De Nittis \textnormal{and} Nicola Gatti\\ ~ Politecnico di Milano\\ ~ Piazza Leonardo da Vinci, 32\\ ~ Milano, Italy\\ ~ \{andrea.celli, giuseppe.denittis, nicola.gatti\}@polimi.it}
\maketitle

\begin{abstract}

The \emph{Team-maxmin equilibrium} prescribes the optimal strategies for a team of rational players sharing the same goal and without the capability of correlating their strategies in strategic games against an adversary. This solution concept can capture situations in which an agent controls multiple resources---corresponding to the team members---that cannot communicate. It is known that such equilibrium always exists and it is unique (unless degeneracy) and these properties make it a credible solution concept to be used in real--world applications, especially in security scenarios. Nevertheless, to the best of our knowledge, the Team--maxmin equilibrium is almost completely unexplored in the literature. In this paper, we investigate bounds of (in)efficiency of the Team--maxmin equilibrium w.r.t. the Nash equilibria and w.r.t. the Maxmin equilibrium when the team members can play correlated strategies. Furthermore, we study a number of algorithms to find and/or approximate an equilibrium, discussing their theoretical guarantees and evaluating their performance by using a standard testbed of game instances.
\end{abstract}

\section{Introduction}

The computational study of game--theoretic solutions concepts is among the most important challenges addressed in the last decade of Computer Science~\cite{deng2002complexity}. These problems acquired particular relevance in Artificial Intelligence, where the goal is to design physical or software agents that must behave optimally in strategic situations. In addition to the well--known Nash equilibrium~\cite{nash1951non}, other solution concepts received attention in the Artificial Intelligence literature  thanks to their application in security domains. Examples include Maxmin equilibrium for zero--sum games under various forms of constraints over the actions of the players~\cite{jain2010security} and Stackelberg (a.k.a. leader--follower) equilibrium~\cite{conitzer2006computing}.

While a large part of the literature focuses on 2--player games, few results are known about games with more players---except for games with a very specific structure, e.g., congestion games~\cite{nisan2007algorithmic}. In this paper, we focus on the Team--maxmin equilibrium proposed by~\cite{von1997team}. It applies to zero--sum games between a team and an adversary. The team is defined as a set of players with the same utility function $U_T$ and without the capability of synchronizing their actions. The adversary is a single player with utility function $-U_T$. These games can model many realistic security scenarios, for example those where multiple non--coordinating agents share the common objective of defending an environment against a malicious attacker. In~\cite{JiangProcaccia2013}, a security setting of such type is studied and an analysis of the price of mis--coordination in the specific proposed security games is conducted. The Team--maxmin equilibrium plays a crucial role also in infinitely repeated games and role assignment problems~\cite{moon2016}, where it is necessary to compute threat points.
The current approach to tackle these problems, in games with more than two players, is considering the correlated threat point~\cite{kontogiannis2008} or employing approximating algorithms that avoid the use of linear programming~\cite{andersen2013}. Our techniques allow the computation punishment strategies (leading to the threat points) in the general scenario in which players, other than the defector, cannot coordinate strategy execution.

The study of Team--maxmin equilibrium is almost completely unexplored. It is known that it always exists, it is unique except for degeneracies, and it is the best Nash equilibrium for the team, but, to the best of our knowledge, only two computational works deal with this solution concept. \cite{DBLP:journals/geb/BorgsCIKMP10} show that the Minmax value (equivalently the Team--maxmin value) is inapproximable in additive sense within $\frac{3}{m^2}$ even in 3--player games with $m$ actions per player and binary payoffs (but nothing is known about the membership to \textsf{APX} class or some super class); \cite{DBLP:conf/wine/HansenHMS08} strengthen the previous complexity result and provide a quasi--polynomial time $\epsilon$--approximation (in additive sense) algorithm. Only~\cite{lim1997rendezvous,alpern1998symmetric} deal with the mathematical derivation for a specific class of games with an adversary, i.e., rendezvous--evasion games. Instead, a number of works deal with team games without adversary. We just cite a few for the sake of completeness. Team games were first proposed in~\cite{palfrey1983strategic} as voting games, then studied in repeated and absorbing games to understand the interaction among the players~\cite{bornstein1994effect,bornstein1996experimental,bornstein1997cooperation,solan2000absorbing} and more recently in Markov games with noisy payoffs~\cite{wang2002reinforcement}.

\textbf{Original contributions} We provide two main contributions. First, we study the relationship, in terms of efficiency for the team, between Nash equilibrium (i.e., when players are not teammates), Team--maxmin equilibrium, and Correlated--team maxmin equilibrium (i.e., the Maxmin equilibrium when all the team members can play in correlated strategies and then can synchronize the execution of their actions). We show that, even in the same instances with binary payoffs, the worst Nash equilibrium may be arbitrarily worse than the Team--maxmin equilibrium that, in its turn, may be arbitrarily worse (in this case only asymptotically) than the Correlated--team maxmin equilibrium. We provide exact bounds for the inefficiency and we design an algorithm that, given a correlated strategy of the team, returns in polynomial time a mixed strategy of the team minimizing the worst--case ratio  between the utility given by the correlated strategy and the utility given by the mixed strategy. Second, we provide some algorithms to find and/or approximate the Team--maxmin equilibrium, we discuss their theoretical guarantees and evaluate them in practice by means of a standard testbed~\cite{nudelman2004run}. We also identify the limits of such algorithms and discuss which ones are the best to be adopted depending on the instance to be solved. For the sake of presentation, the proofs of the theorems are presented in the Appendices.

\section{Preliminaries}\label{sec:prob_formulation}

A normal--form game is a tuple $(N,A,U)$ where: $N = \{1, 2, \ldots, n\}$ is the set of players; $A = \bigtimes_{i \in N} A_i$ is the set of player~$i$'s actions, where $A_i=\{a_1,a_2,\ldots,a_{m_i}\}$; $U = \{U_1, U_2, \ldots, U_n\}$ is the utility function of player~$i$, where $U_i: A \rightarrow \mathbb{R}$. A strategy profile is defined as $s = (s_1, s_2, \ldots, s_n)$, where $s_i \in \Delta(A_i)$ is player~$i$'s mixed strategy and $\Delta(A_i)$ is the set of all the probability distributions over $A_i$. As customary, $-i$ denotes the set containing all the players except player~$i$. We study games in which the set of players $T = \{1, 2, \ldots, n-1\}$ constitutes a team whose members have the same utility function $U_T$. Player~$n$ is an adversary of the team and her utility function is $-U_T$.

When the teammates cannot coordinate at all and therefore no player can communicate with the others, and each player takes decisions independently, the appropriate solution concept is the Nash equilibrium, which prescribes a strategy profile where each player~$i$'s strategy $s_i$ is a best response to $s_{-i}$. In 2--player zero--sum games, a Nash equilibrium is a pair of Maxmin/Minmax strategies and can be computed in polynomial time. In arbitrary games, the computation of a Nash equilibrium is \textsf{PPAD}--complete even when the number of players is fixed~\cite{daskalakis2009complexity}. Instead, when the teammates can coordinate themselves, we distinguish two forms of coordinations: \emph{correlated}, in which a correlating device decides a joint action (i.e., an action profile specifying one action per teammate) and then communicates each teammate her action, and \emph{non--correlated}, in which each player plays independently from the others.

When the coordination is non--correlated, players are subject to the inability of correlating their actions, and their strategy $s_i$ is mixed, as defined above for a generic normal--form game. In other words, teammates can jointly decide their strategies, but they cannot synchronize their actions, which must then be drawn independently. 
The appropriate solution concept for such setting is the Team--maxmin equilibrium. 

A Team--maxmin equilibrium is a Nash equilibrium with the properties of being unique (except for degeneracies) and the best one for the team. These property are very appealing in real--world settings, since they allow to avoid the equilibrium selection problem which affects the Nash equilibrium. In security applications, for instance, the equilibrium uniqueness allows to perfectly forecast the behavior of the attacker (adversary). When the number of players is given, finding a Team--maxmin equilibrium is \textsf{FNP}--hard and the Team--maxmin value is inapproximable in additive sense even when the payoffs are binary~\cite{DBLP:conf/wine/HansenHMS08}. \footnote{Rigorously speaking, \cite{DBLP:conf/wine/HansenHMS08} studies Minmax strategy when there is a single max player and multiple min players. The problem of finding the Team--maxmin equilibrium in zero--sum adversarial team games can be formulated as the problem of finding such Minmax strategy and \emph{vice versa}.} In~\cite{DBLP:conf/wine/HansenHMS08}, the authors provide a quasi--polynomial--time $\epsilon$--approximation (in additive sense) algorithm. Furthermore, a Team--maxmin equilibrium may contain irrational probabilities even with 2 teammates and 3 different values of payoffs.\footnote{The proof, provided in~\cite{DBLP:conf/wine/HansenHMS08}, contains a minor flaw. In the Appendices, we provide a correct revision of the proof with all the calculations, omitted in the original proof.} It is not known any experimental evaluation of algorithms for finding the Team--maxmin equilibrium.

When players can synchronize their actions, the team strategy is said to be correlated. Given the set of team action profiles defined as $A_T = \bigtimes_{i \in T} A_i$, a correlated team strategy is defined as $p \in \Delta(A_T)$. In other words, teammates can jointly decide and execute their strategy. The team is then equivalent to a single player whose actions are joint team action profiles. In such case, the appropriate solution concept for the team and the adversary is a pair of Maxmin/Minmax strategies that, for the sake of clarity, we call in this paper \emph{Correlated--team maxmin equilibrium}. This equilibrium can be found by means of linear programming since it can be formulated as a maxmin problem in which the max player's action space is given by the Cartesian product of the action space of each teammate.
Notice that the size of the input is exponential in the number of teammates and therefore approximation algorithms for games with many team members are necessary in practice.

Furthermore, it is not known the price---in terms of inefficiency---paid by a team due to the inability of synchronizing the execution of their actions. This would allow to understand how the Team--maxmin equilibrium is inefficient w.r.t. the Correlated--team maxmin equilibrium, or equivalently, how well the Team–-maxmin equilibrium approximates the Correlated–-team maxmin equilibrium. Another open problem is studying the gain a set of players sharing the same goal would have in forming a team and coordinating their mixed strategies (i.e., how is the Nash equilibrium inefficient w.r.t. the Team–-maxmin equilibrium, or equivalently, how well the Nash equilibrium approximates the Team–-maxmin equilibrium).

\section{Nash, Team-maxmin, and \\Correlated--team maxmin equilibria}

We study the relationships between Nash equilibrium and Team--maxmin equilibrium in terms of efficiency for the team. In our analysis, we resort to the concept of Price of Anarchy (\textsc{PoA}), showing that Nash equilibrium---precisely, the worst case Nash equilibrium---may be arbitrarily inefficient w.r.t. the Team--maxmin equilibrium---corresponding to the best Nash equilibrium for the team. In this case the \textsc{PoA} provides a measure about the inefficiency that a group of players with the same goal would have if they do not form a team. To have coherent results with the definition of \textsc{PoA}, we consider games with payoffs in the range $[0,1]$. We observe that our results will hold without loss of generality since, given any arbitrary game, we can produce an equivalent game in which the payoffs are in such a range by using an affine transformation. Furthermore, for the sake of presentation, we consider only games in which $m_1=\ldots = m_n=m$. The generalization of our results when players may have a different number of actions is straightforward.

\begin{theorem} The Price of Anarchy (\textsc{PoA}) of the Nash equilibrium w.r.t. the Team--maxmin equilibrium may be \textsc{PoA}$ =\infty$ even in games with 3 players (2 teammates), 2 actions per player, and binary payoffs.
\label{thm:POA}
\end{theorem}

In order to evaluate the inefficiency of the Team--maxmin equilibrium w.r.t. the Correlated--team maxmin equilibrium, we introduce a new index similar to the {\em mediation value} proposed in~\cite{ashlagi2008value} and following the same rationale of the \textsc{PoA}. We call such an index Price of Uncorrelation (\textsc{PoU}) and we define it as the ratio between the team's utility provided by the Correlated--team maxmin equilibrium and that one provided by the Team--maxmin equilibrium. \textsc{PoU} provides a measure of the inefficiency due to the impossibility, for the teammates, of synchronizing the execution of their strategies.

\begin{definition}
Let us consider an $n$--player game. The Price of Uncorrelation (\textsc{PoU}) is defined as \textsc{PoU} $= \frac{v^{\text{team}}_{\text{C}}}{v^{\text{team}}_{\text{M}}}\geq 1$
where $v^{\text{team}}_{\text{C}}$ is Correlated--team maxmin value of the team and $v^{\text{team}}_{\text{M}}$ is Team--maxmin value of the team.
\end{definition}

We initially provide a lower bound over the worst--case \textsc{PoU}.

\begin{theorem}
The \textsc{PoU} of the Team--maxmin equilibrium w.r.t. the Correlated--team maxmin equilibrium may be \textsc{PoU}$ =m^{n-2}$ even in games with binary payoffs.
\label{thm:POU}
\end{theorem}

Now, we provide an upper bound over the worst--case  \textsc{PoU}.

\begin{theorem}
Given any $n$--player game and a Correlated--team maxmin equilibrium with a utility of $v$ for the team, it is always possible to find in polynomial time a mixed strategy profile for the team providing a utility of at least $\frac{v}{m^{n-2}}$ to the team and therefore \textsc{PoU} is never larger than $m^{n-2}$.\label{thm:recostruction}
\end{theorem}

We observe that Theorem~\ref{thm:POU} shows that the upper bound of \textsc{PoU} is at least $m^{n-2}$, while Theorem~\ref{thm:recostruction} shows that \textsc{PoU} cannot be larger than $m^{n-2}$. Therefore, \textsc{PoU} is arbitrarily large only asymptotically. In other words, \textsc{PoU} $ =\infty$ only when $m$ or  $n$ go to $\infty$.\footnote{A more accurate bound can be obtained by substituting $m$ with the size of the equilibrium support, showing that the inefficiency increases as the equilibrium support increases.} More importantly, the proof of Theorem~\ref{thm:recostruction} provides a polynomial--time algorithm to find a mixed strategy of the team given a correlated strategy and this algorithm is the best possible algorithm in terms of worst--case minimization of \textsc{PoU}. 
The algorithm is simple and computes mixed strategies for the team members as follows. Given the Correlated--team maxmin equilibrium $p \in \Delta(A_1\times \ldots \times A_{n-1})$, the mixed strategy of player $1$ $(s_1)$ is such that each action $a_1$ is played with the probability that $a_1$ is chosen in $p$, that is $s_1(a_1)=\sum_{a_{-1}\in A_{-1}} p(a_{1},a_{-1})$. Every other team member $i \in N\setminus\{1,n\}$ plays uniformly over the actions she plays with strictly positive probability in $p$. Since the computation of a Correlated--team maxmin equilibrium can be done in polynomial time, such an algorithm is a polynomial--time approximation algorithm for the Team--maxmin equilibrium.

Furthermore, notice that \textsc{PoU} rises polynomially in the number of actions~$m$ and exponentially in the number of players~$n$. Interestingly, the instances used in the proof of Theorem~\ref{thm:POU} generalize the instances used in the proof of Theorem~\ref{thm:POA}. Indeed, it can be observed that the \textsc{PoA} of the Nash equilibrium w.r.t. the Team--maxmin equilibrium is $\infty$ in the instances used in the proof of Theorem~\ref{thm:POU}. Therefore, there are instances in which the worst Nash equilibrium is arbitrarily worse than the Team--maxmin equilibrium and, in its turn, the Team--maxmin equilibrium is  arbitrarily worse (in this case only asymptotically) than the Correlated--team maxmin equilibrium. 

For the sake of completeness, we state the following result, showing the lower bound of \textsc{PoU}.

\begin{theorem}
The \textsc{PoU} of the Team--maxmin equilibrium w.r.t. the Correlated--team maxmin equilibrium may be \textsc{PoU}$ =1$ even in games with binary payoffs.
\label{thm:POUminimo}
\end{theorem}

\section{Algorithms to find and/or approximate a Team-maxmin equilibrium }\label{sec:prob_analysis}

In the following, we describe four algorithms to find/approximate the Team--maxmin equilibrium.

\subsubsection{Global optimization}
The problem of finding the Team--maxmin equilibrium can be formulated as a non--linear non--convex mathematical program as follows:

\begin{scriptsize}
\begin{align*}
\max\limits_{v,s_i} \quad	&		v 																						\\
\text{s.t.}\quad	&		v - \sum\limits_{a_T \in A_T} U_T (a_T, a_n) \prod\limits_{i \in T}s_i(a_i) \leq 0 		& 	\forall a_n \in A_n		\\
			&		\sum\limits_{a_i \in A_i} s_i(a_i) = 1										&	\forall i \in T			\\
			&		s_i(a_i) \ge 0 														&	\forall i \in T, a_i \in A_i
\end{align*}
\end{scriptsize}

\noindent In order to find an exact (within a given accuracy) Team--maxmin equilibrium, we resort to global optimization tools. Global optimization obviously requires exponential time. In particular, we use BARON~\cite{ts:05}  solver, since it is the best performing solver for completely continuous problems among all the existing global optimization solvers~\cite{neumaier2005comparison}. Most importantly, BARON, if terminated prematurely, returns a lower bound, corresponding to the value of the best solution found so far, and an upper bound, corresponding to the tightest upper bound over the Team--maxmin value found so far. 

\subsubsection{Reconstruction from correlated strategies} 
We approximate the Team--maxmin equilibrium by using a simple variation of the algorithm described previously to find a mixed strategy from a correlated one. First, the algorithm finds a Correlated--team maxmin by means of linear programming. Second, we derive the mixed strategy. The algorithm can be parametrized by exchanging player~1 with each player of the team. This leads to $n-1$ different mixed strategies from the same correlated strategy.  The algorithm returns the best one for the team. Since the Correlated--team maxmin equilibrium is always better than the Team--maxmin equilibrium, this algorithm assures an approximation factor of at least $\frac{1}{m^{n-2}}$ showing that the problem is in \textsf{Poly--APX} when $n$ is given.

\subsubsection{Support enumeration}\label{subsec:miltersen}
In~\cite{DBLP:conf/wine/HansenHMS08}, the authors show how in $n$--players finite strategic games the minmax value of a player can be approximated (from above) within an arbitrary additive error $\epsilon > 0$. The algorithmic approach to guarantee such approximation leverages the concept of {\em simple} strategies as introduced in~\cite{lipton1994simple} and can be exploited to approximate the Team--maxmin value, as we fully report in Algorithm~\ref{alg:miltersen}.

\begin{algorithm}[!t]
\caption{\texttt{SupportEnumeration}}
\begin{scriptsize}
\begin{algorithmic}[1]
\State $v^* = +\infty$ \label{alg:init}
\ForAll{$i \in T$} 
	\State $P_i = \{(V_i^1, m_i^1), (V_i^2, m_i^2), \ldots \mid \forall j, V^j_i \subseteq A_i, \sum\limits_{a \in V_i}m_i^j(a) = \Gamma \}$\label{alg:Pis_def}
\EndFor
\State $C = \bigtimes_{i \in T} P_i$ \label{alg:cartesian}
\ForAll{$\big((V_1, m_1), (V_2,m_2), \ldots, (V_{n-1},m_{n-1}) \big) \in C$} \label{alg:enum}
\ForAll{ $i \in T$}
\State $s_i (a_i) =   \begin{cases}
    \frac{m_i(a_i)}{\Gamma}, & \text{if } a_i \in V_i \\
    0 & \text{otherwise}
  \end{cases}$ \label{alg:uniform_strat}
\EndFor
\State $v^* = \max \{ v^*, \min\limits_{s_n} U_T(s_1, s_2, \ldots, s_{n-1})\}$ \label{alg:minimizer}
\EndFor
\State \textbf{return} $v^*$
\end{algorithmic}
\end{scriptsize}
\label{alg:miltersen}
\end{algorithm}

The algorithm enumerates joint action multi--sets (specifying, for each player $i$, a subset of actions that can contain duplicate elements) of cardinality $\Gamma = \big\lceil \frac{\ln |A_i|}{2\epsilon^2} \big\rceil$. The strategy for each player is then obtained from an uniform distribution over the considered multi--set (for example, if action $a_i$ has $k$ duplicates it will be selected with probability $k/\Gamma$). The adversary's best response and the corresponding value for the team is then computed. The algorithm returns the joint support maximizing the value of the team.  With a slight adaptation of the analysis made in~\cite{DBLP:conf/wine/HansenHMS08}, it can be easily shown that Algorithm~\ref{alg:miltersen} approximates the Team--maxmin value with additive error of at most $\epsilon$ with a number of iterations equal to ${m + \Gamma - 1 \choose \Gamma}^{n-1}$.

In the table below we report, for some $m$, the number of iterations required by the algorithm to assure a given approximation with additive error not larger than $\epsilon$.

\begin{scriptsize}
\begin{center}
\begin{tabular}{|l||r|r|r|r|r|r|}
\hline
$m$ 						&	5		&	5		&	5		&	10		&	10		&	10		\\	\hline
$\epsilon$ 		&	$0.9$	&	$0.5$	&	$0.1$	&	$0.9$	&	$0.5$	&	$0.1$	\\	\hline
iterations					&	$> 2^{4}$	&	$>2^{12}$	&	$>2^{41}$	&	$>2^{11}$	&	$>2^{21}$ 	&	$>2^{87}$ \\	\hline
\end{tabular}
\end{center}
\end{scriptsize}

\noindent As it can be seen, the algorithm can provide only a coarse guarantee (in additive sense) in small games, requiring however a large number of iterations.

\subsubsection{Iterating linear programming}\label{subsec:iterate_lp}

In Algorithm~\ref{alg:iterated-lp} we propose a method we call \texttt{IteratedLP} based on solving iteratively a Maxmin problem between 2 players (a member of the team and the adversary) by linear programming.
\begin{algorithm}[!t]
\caption{\texttt{IteratedLP}}
\begin{scriptsize}
\begin{algorithmic}[1]
\State $\forall i \in T$,  $s^{cur}_i \leftarrow \hat{s}_{i}$\label{alg2:init}
\Repeat
\ForAll {$i \in T$} \label{alg2:multiLP}
\State 
$
\begin{array}{ll}
v_i^*=\max & v_i																								\\
v_i \leq \sum\limits_{a_T \in A_T} x_{a_i} U_T (a_T, a_n) \prod\limits_{j \in T \setminus \{i\}}s^{cur}_j(a_j) & \forall a_n \in A_n		\\
\sum\limits_{a_i \in A_i}x_{a_i}=1							& \forall a_i \in A_i: \\ & x_{a_i} \ge 0
\end{array}
$
\label{alg2:lp}
\EndFor
\State $i' = \arg\max\limits_{i \in T} v^*_i$ \label{alg2:select_best_lp}
\State
$s^{cur}_i (a_i) =   \begin{cases}
    x^*_{a_i} & \text{if } i = i' \\
    s^{cur}_i (a_i) & \text{otherwise}
  \end{cases}$\label{alg2:update}
\Until {convergence \textbf{or} timeout}\label{alg2:terminate}
\State \textbf{return} $s^{cur}$
\end{algorithmic}
\end{scriptsize}
\label{alg:iterated-lp}
\end{algorithm}

It works by maintaining a current solution $s^{cur}$ which specifies a strategy for each team member. It is initialized (Line~\ref{alg2:init}) with a starting solution $\hat{s}$ which, in principle, can prescribe an arbitrary set of strategies for the team (e.g., uniform randomizations). Then for each team member $i$ (Line~\ref{alg2:multiLP}), we instantiate and solve the specified linear program (Line~\ref{alg2:lp}). The decision variables of this LP are $v_i$ and, for each action $a_i$ of player $i$, $x_{a_i}$. We maximize $v_i$ subject to the upper bound given by the first constraint, where we assumed that the strategy of player $i$ (relabeled with $x$ to distinguish it) is a variable while the strategies of the other team members are constants, set to the associated value specified by the current solution. (Notice that, in the LP, $a_j$ is the action that team member $j$ plays in the team action profile $a_T$). The optimal solution of the LP is given by $v_i^*$ and $x^*$, representing the Maxmin strategy of team member $i$ once the strategies of teammates have been fixed to the current solution. Once the LP has been solved for each $i$, the algorithm updates the current solution in the following way (Line~\ref{alg2:update}): the strategy of the team member that obtained the best LP optimal solution is replaced with the corresponding strategy from the LP. This process continuously iterates until convergence or until some timeout is met. At each iteration of the algorithm, the value of the game increases (non--strictly) monotonically. We run it using multiple random restarts, i.e., generating a set of different initial assignments $\hat{s}$ (Line~\ref{alg2:init}). Once convergence is  achieved, we pass to the next random restart generating a new strategy profile for the team. 

A crucial question is whether there are initializations that are better or worse than others. We can prove the following.

\begin{theorem}
When the algorithm is initialized with a uniform strategy for every player, the worst--case approximation factor of Algorithm~\ref{alg:iterated-lp} is at least $\frac{1}{m^{n-1}}$ and at most $\frac{1}{m^{n-2}}$. When instead the algorithm is initialized with a pure strategy, the worst--case approximation factor is 0.
\end{theorem}

We leave open the problem of studying how the worst--case approximation factor varies for other initializations.

\section{Experimental evaluation}
\label{sec:experiments}

\subsection{Experimental setting}
\label{subsec:testbed}
Our experimental setting is based on instances of \texttt{RandomGames} class generated by GAMUT~\cite{nudelman2004run}. Specifically, once a game instance is generated, we extract the utility function of player~$1$ and assign it to all the team members. Furthermore, in each generated game instance, the payoffs are between 0 and 1. We use 100 game instances for each combination of $n$ and $m$ where $n \in \{3,4,5\}$ and $m$ is as follows:

\begin{scriptsize}
\begin{equation*}
m = \begin{cases}
\begin{array}{lll}
5\;\text{to}\;40, &\text{step} = 5, & n = 3\\
50\;\text{to}\;150, &\text{step} = 10, & n = 3\\
5\;\text{to}\;50, &\text{step} = 5, & n = 4\\
5\;\text{to}\;30, &\text{step} = 5, & n = 5 
\end{array}
\end{cases}.
\end{equation*}
\end{scriptsize}

\noindent Algorithms are implemented in Python 2.7.6, adopting GUROBI 6.5.0~\cite{gurobi} for linear mathematical programs, AMPL 20160310~\cite{Fourer:1989:AMP:107479.107491} and BARON 14.4.0~\cite{ts:05,sahinidis:baron:14.4.0} for global optimization programs. We set a timeout of 60 minutes for the resolution of each instance. All the algorithms are executed on a UNIX computer with 2.33GHz CPU and 128 GB RAM.

\subsection{Experimental results}
\label{subsec:algorithms_characterization}

\subsubsection{Global optimization} 
\begin{figure}[!htbp]
\centering
\includegraphics[width=0.5\textwidth]{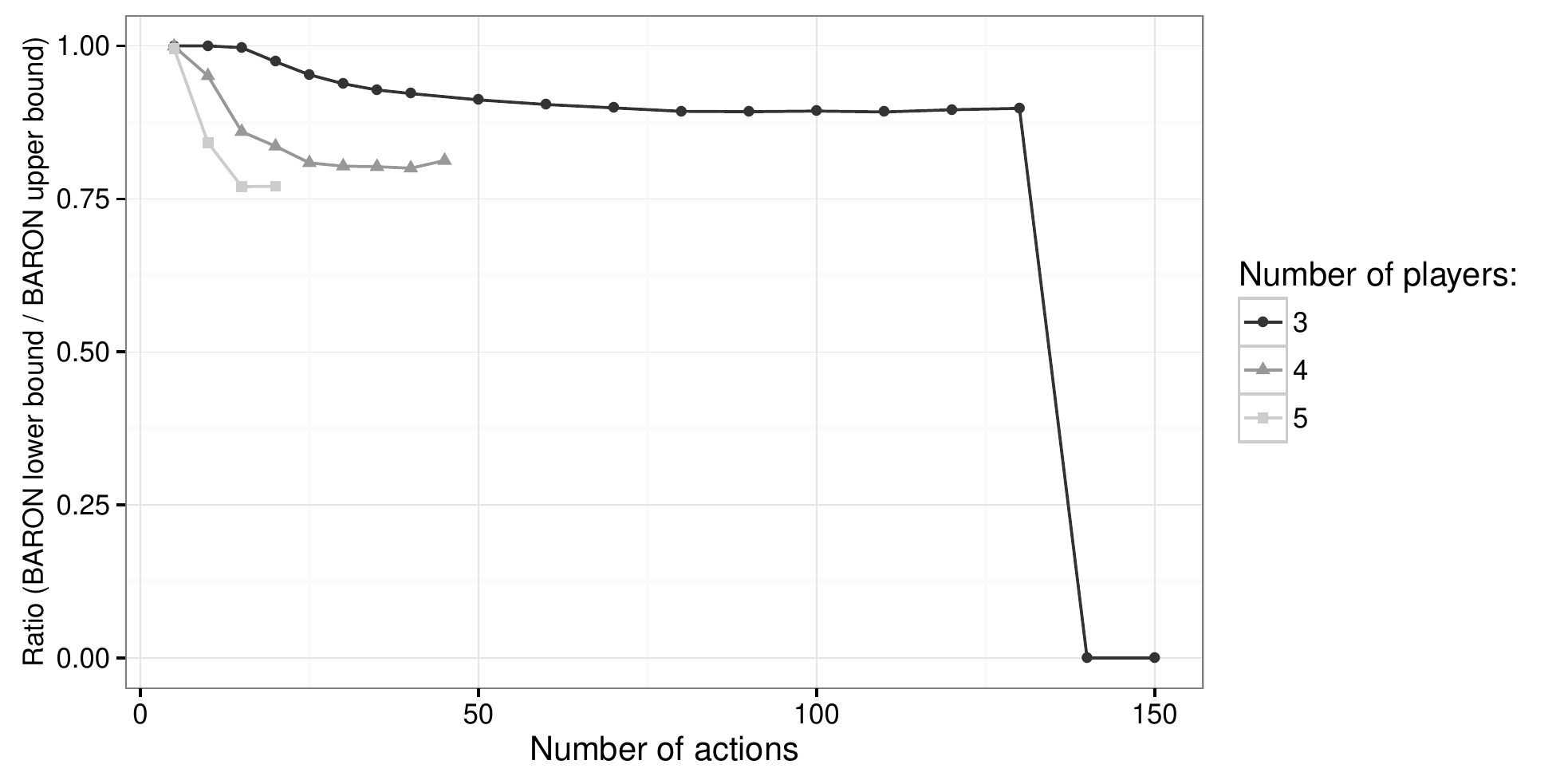}\label{fig:global_opt_mean}
\caption{Average (empiric) approximation ratio (lower bound / upper bound) of the solutions returned by BARON.}
\label{fig:baron}
\end{figure}

The average quality of the solutions is reported in Fig.~\ref{fig:baron} in terms of ratio between the lower (a.k.a. primal) bound and the upper (a.k.a. dual) bound returned by BARON once terminated. When BARON finds the optimal solution (up to an accuracy of $10^{-9}$), the lower bound equals the upper bound achieving a ratio of~1. This happens with $n=3$ up to $m=15$ (except for some outliers), with $n\in\{4,5\}$ and $m=5$. For larger instances with $n=3$ up to $m=130$, with $n=4$ up to $m=45$ and with $n=5$ up to  $m=20$, BARON returns an approximate solution with a ratio always larger than 0.7. With $n=3$ and $m\in\{140,150\}$, BARON returns a very high upper bound such that the ratio is close to zero.  With larger instances, BARON does not run due to memory limits. Hence, BARON demonstrates to be an excellent tool to approximate the Team--maxmin equilibrium especially with $n=3$ (note that, with $n=3$ and $m=130$, the number of outcomes is larger than 2~millions).

\begin{figure*}[!t]
\begin{scriptsize}
\centering

  \subfigure[3--players Reconstruction from correlated strategies.] {\includegraphics[width=0.33\textwidth]{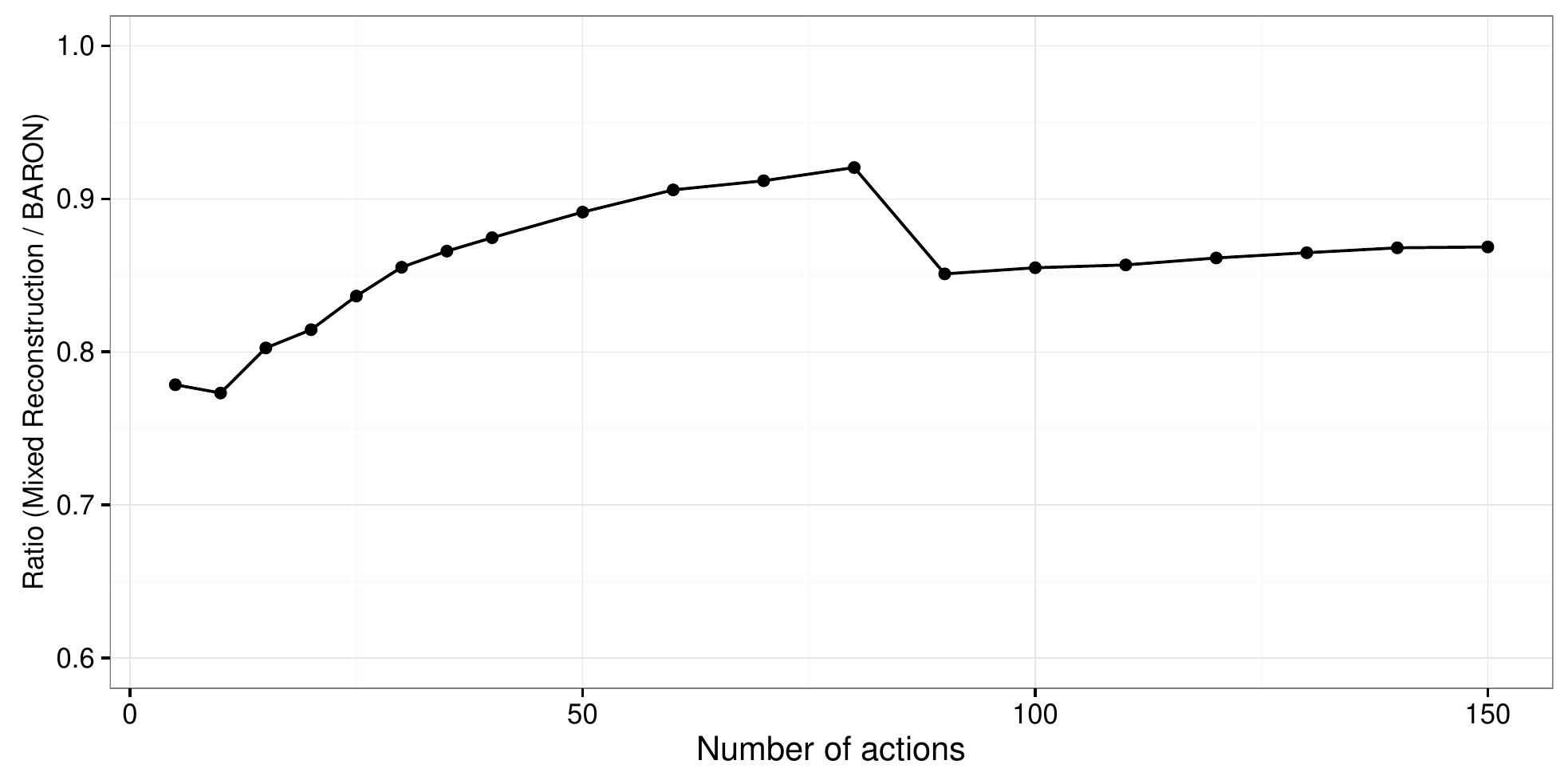}\label{fig:frommixed3pl}}
  \subfigure[3--players Support enumeration.] {\includegraphics[width=0.33\textwidth]{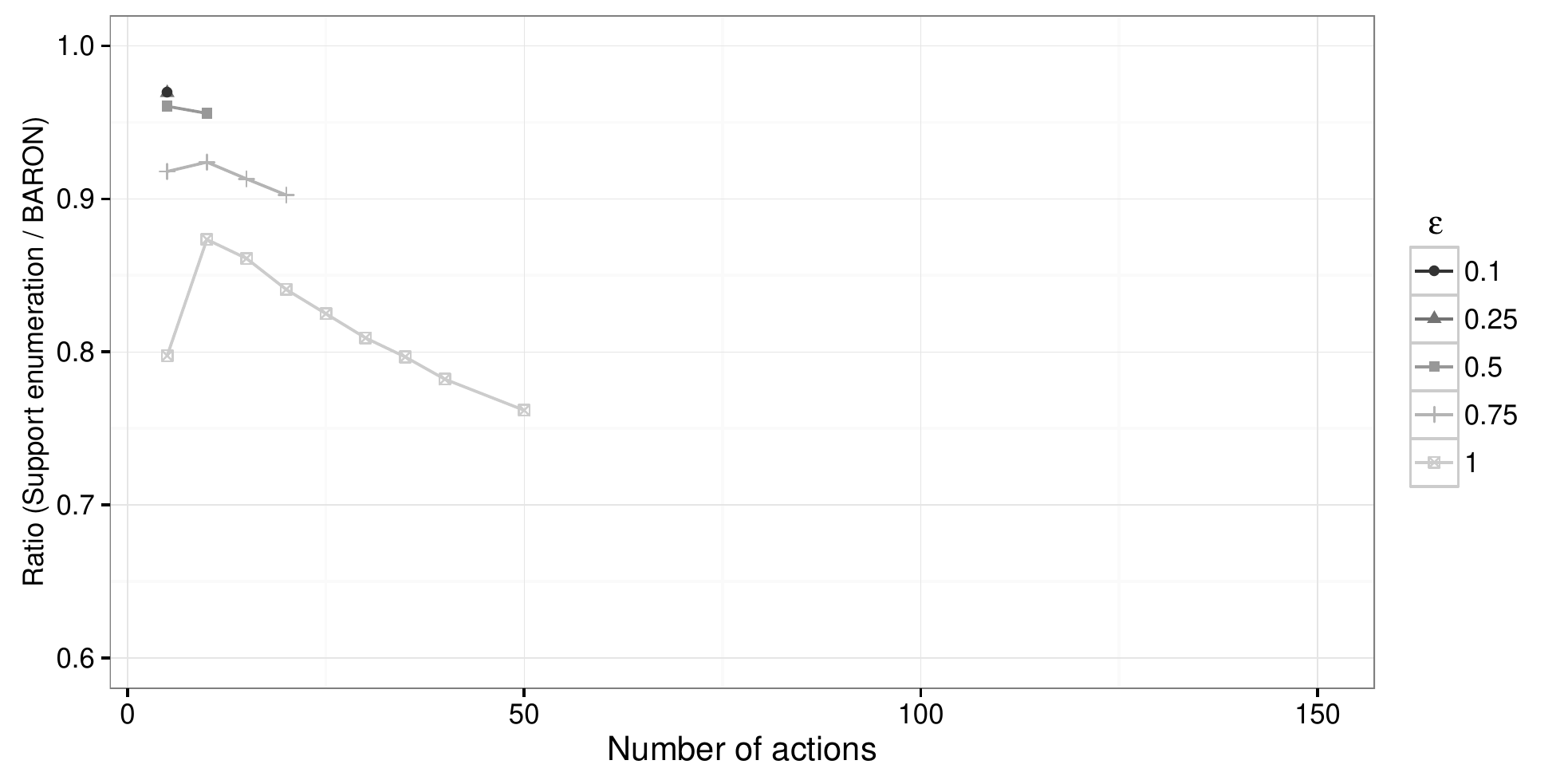}\label{fig:support_enum_3pl}}
  \subfigure[3--players Iterating linear programming.] {\includegraphics[width=0.33\textwidth]{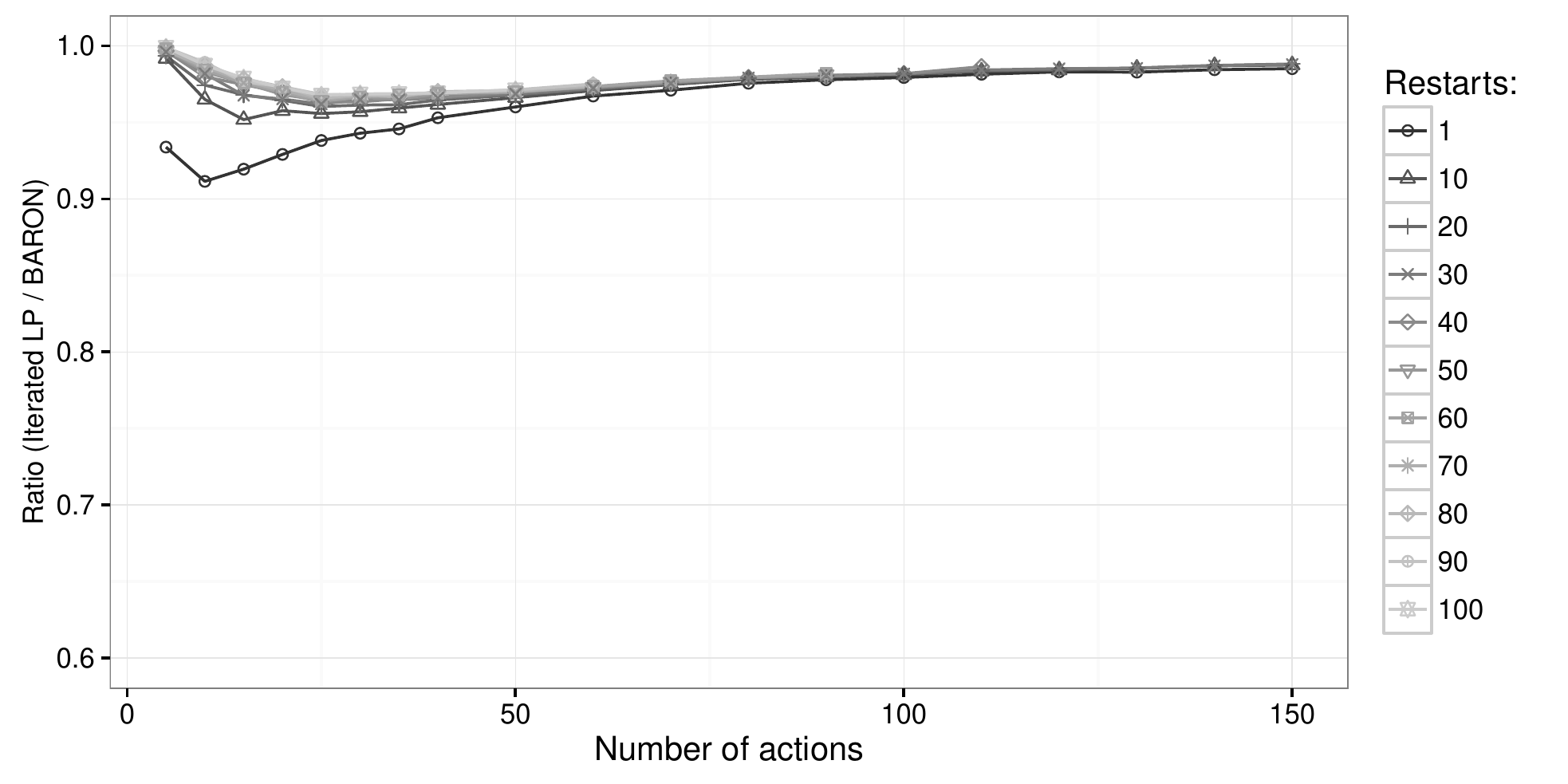}\label{fig:iter_mean_3pl}}

  \subfigure[4--players Reconstruction from correlated strategies.] {\includegraphics[width=0.33\textwidth]{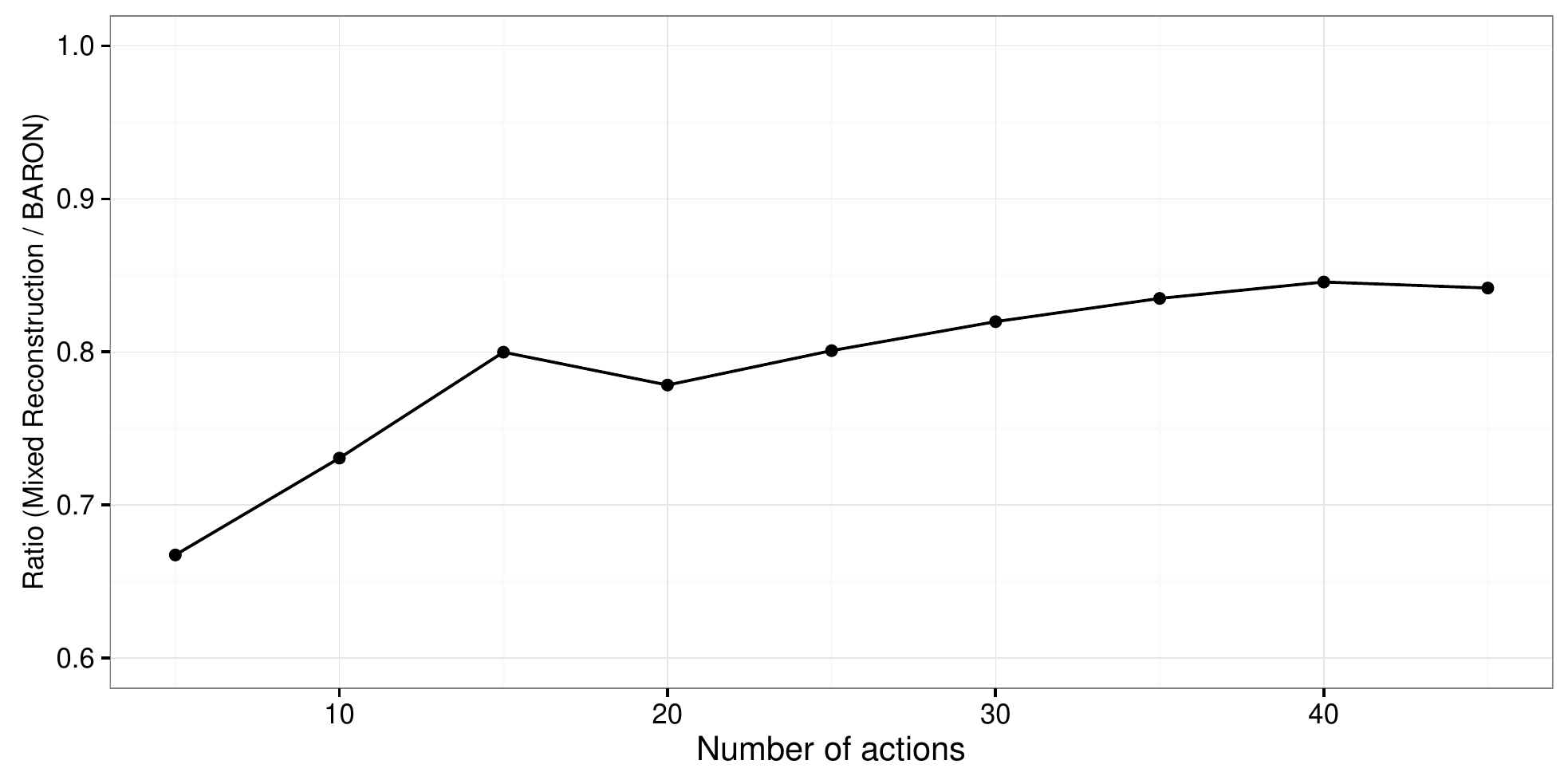}\label{fig:frommixed4pl}}
  \subfigure[4--players Support enumeration.] {\includegraphics[width=0.33\textwidth]{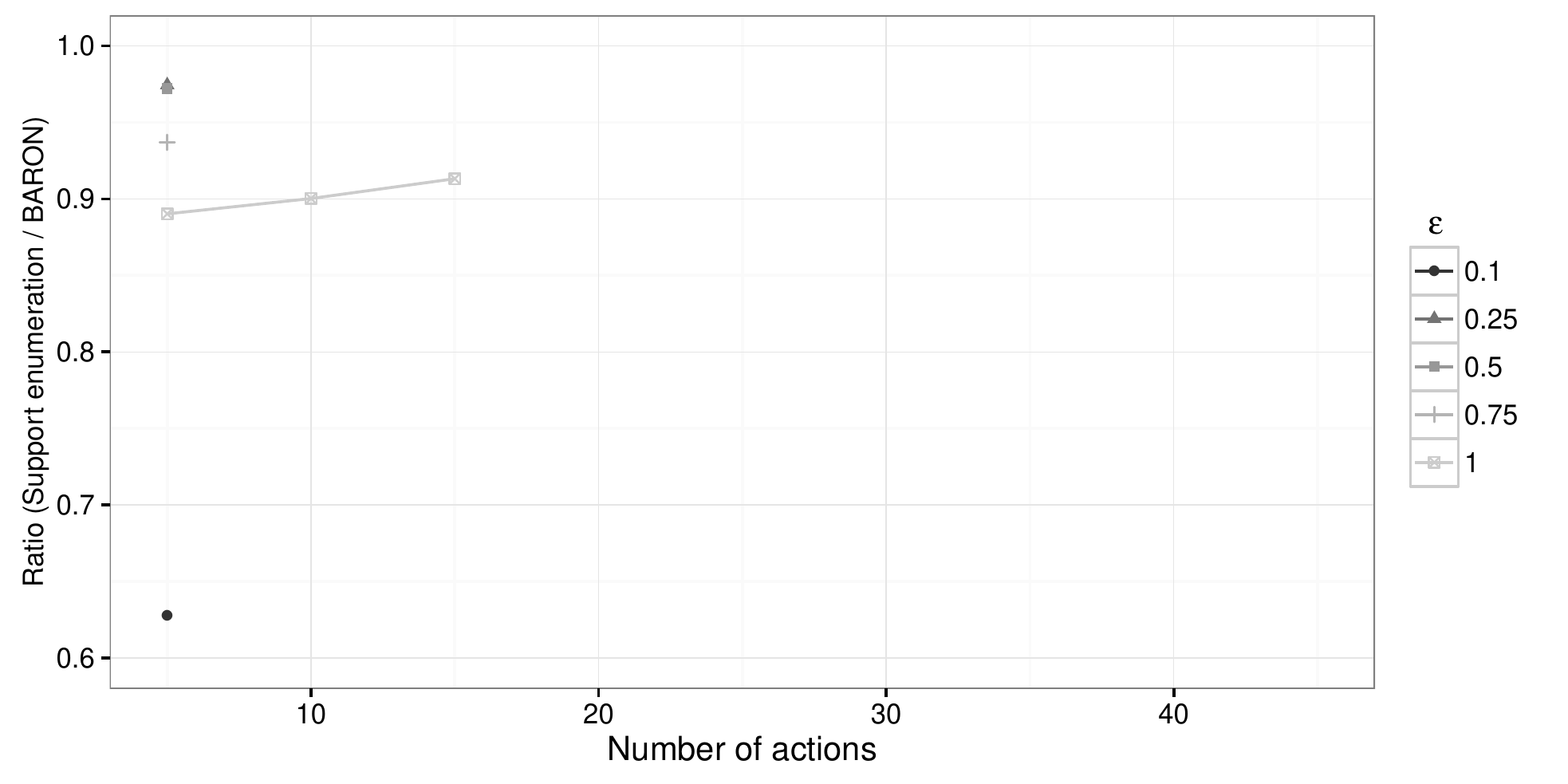}\label{fig:support_enum_4pl}}
  \subfigure[4--players Iterating linear programming.] {\includegraphics[width=0.33\textwidth]{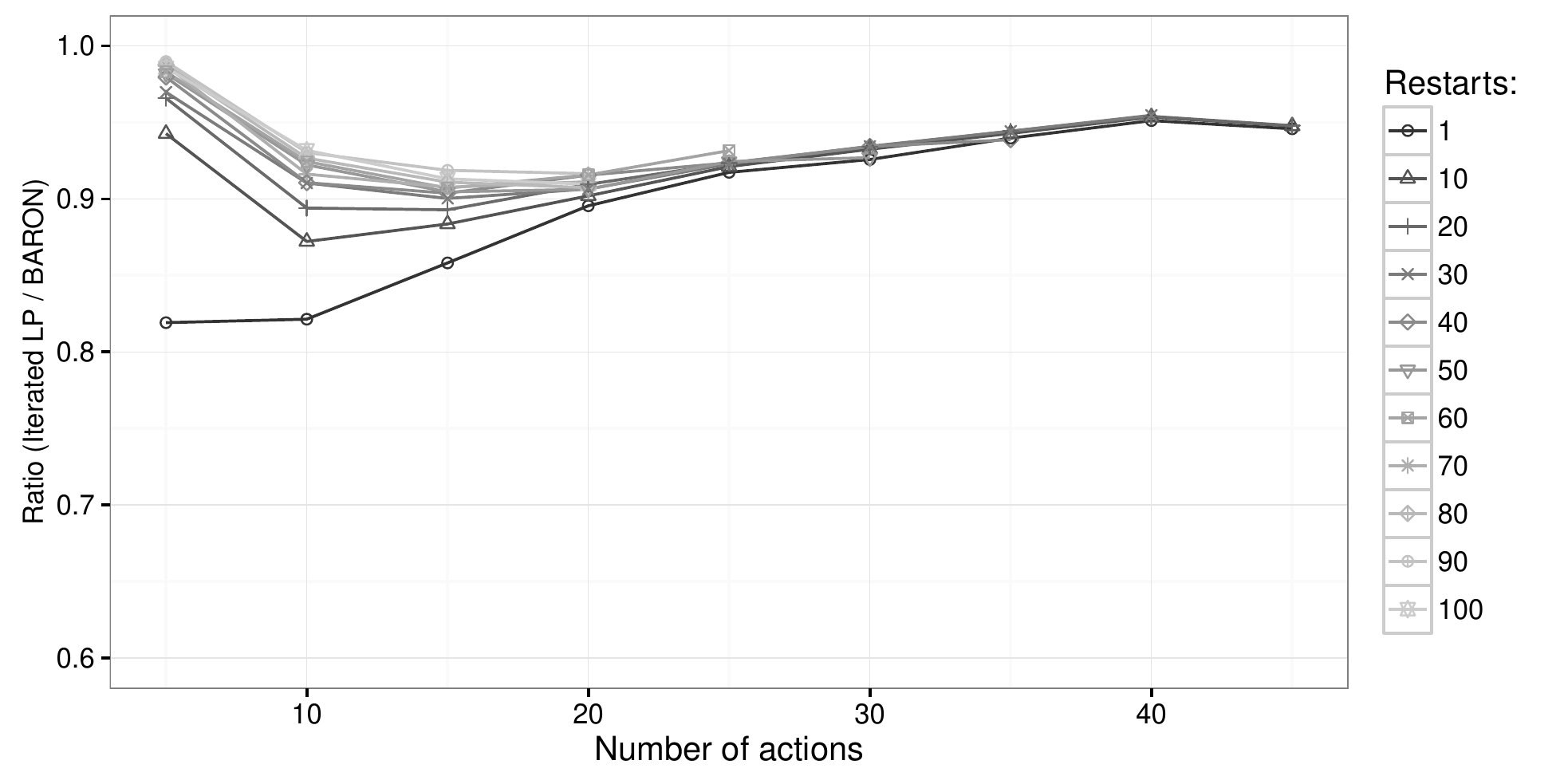}\label{fig:iter_mean_4pl}}

  \subfigure[5--players Reconstruction from correlated strategies.] {\includegraphics[width=0.33\textwidth]{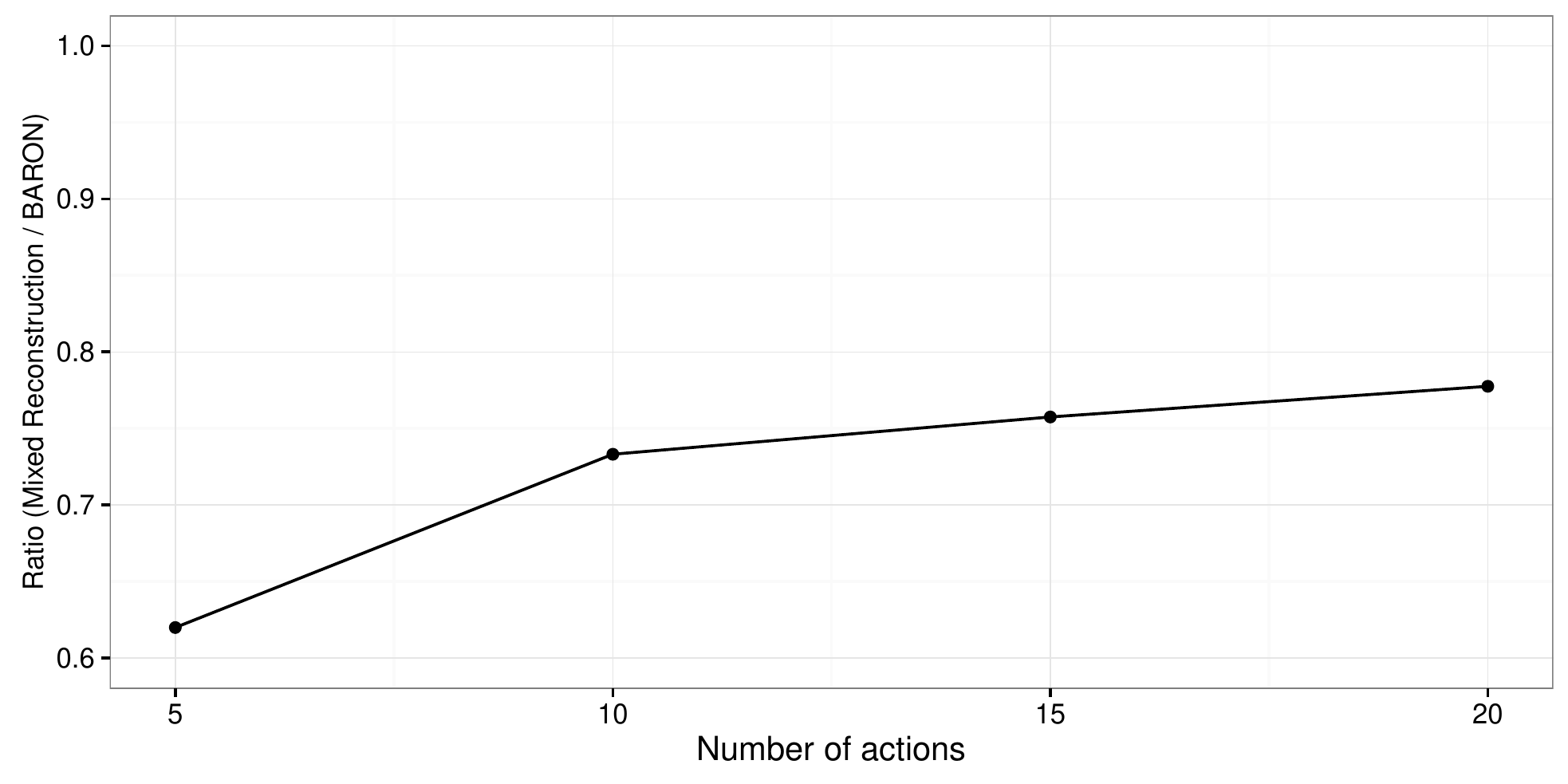}\label{fig:frommixed5pl}}
  \subfigure[5--players Support enumeration.] {\includegraphics[width=0.33\textwidth]{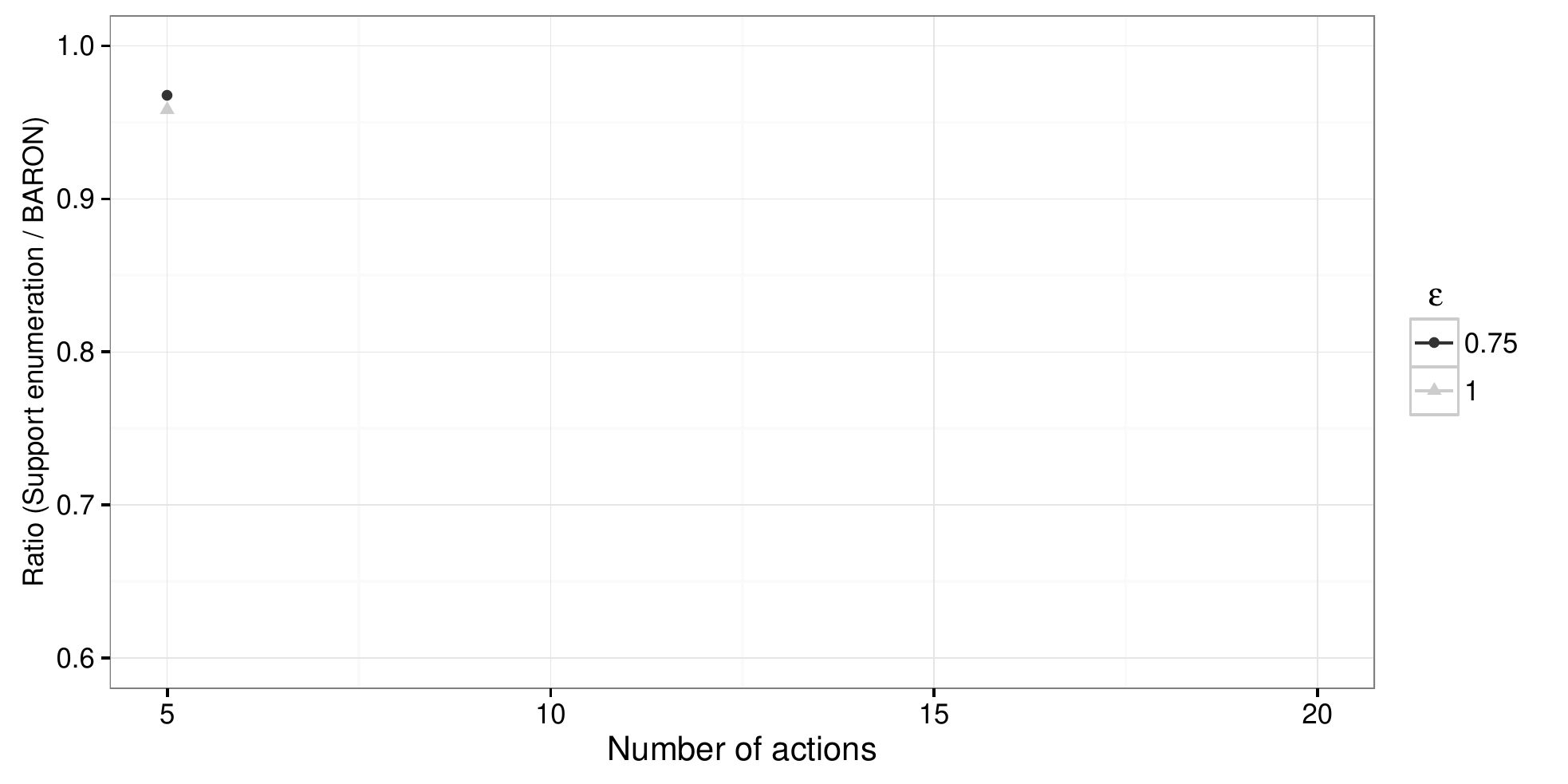}\label{fig:support_enum_5pl}}
  \subfigure[5--players Iterating linear programming.] {\includegraphics[width=0.33\textwidth]{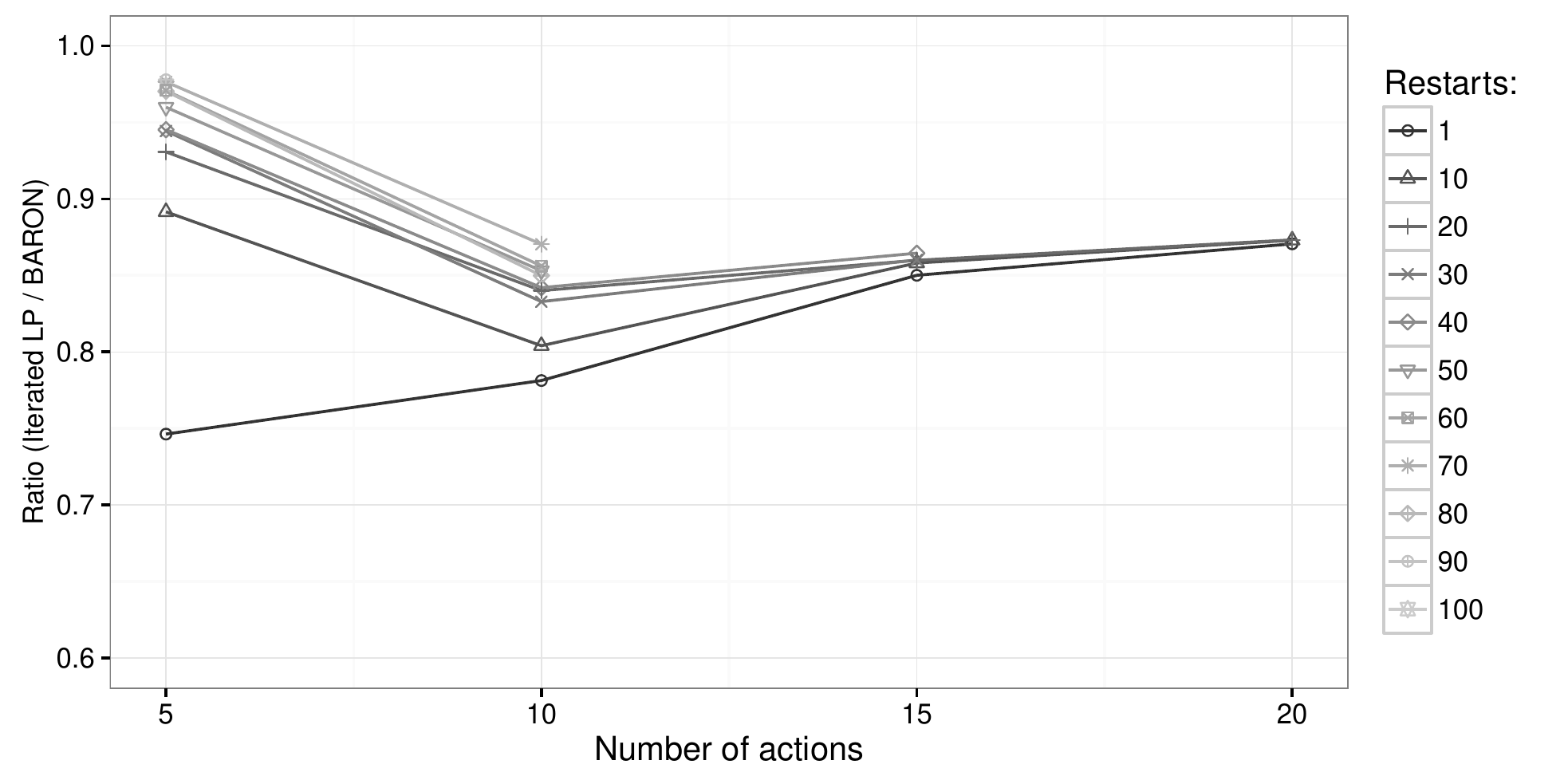}\label{fig:iter_mean_5pl}}

\caption{Average approximation performance of the proposed algorithms w.r.t. global optimization.}\label{fig:iter_supp_enum}
\end{scriptsize}
\end{figure*}

\subsubsection{Other algorithms}

We report in Fig.~\ref{fig:iter_supp_enum} the average performance of the other three algorithms described in the previous section in terms of ratio between team value of the strategy returned by each single algorithm and the lower bound returned by BARON. A ratio smaller than 1 means that the given algorithm provides a solution with a quality worse than the solution returned by BARON. 

Let us focus on the reconstruction from  correlated strategies. This algorithm solves all the instances of our experimental settings, including also the instances that BARON does not solve due to memory limits. However, the quality of the solutions is always worse than the solutions returned by BARON. More precisely, we notice that the ratio is always larger than 0.6 and, with $n=3$, it is larger than 0.8. Hence, the quality of the solutions w.r.t. the upper bound of BARON is always larger than 0.5, thus achieving at least $1/2$ of the Team--maxmin value. As expected, the quality decreases as the number of players increases. Instead, surprisingly, the quality increases as the number of actions per player increases (we provide a motivation for that  below). 

Let us focus on the support enumeration algorithm. We report the performance of the algorithm for $\epsilon\in \{0.10,0.25,0.50,0.75,1.00\}$. 
As expected, the algorithm does not scale. Indeed, when $\epsilon$ is set $<1$, the algorithm terminates only for $m\leq 20$ with $n=3$, and only for $m=5$  with $n \in \{4,5\}$. This shows that the algorithm can be practically applied only when $\epsilon = 1$, but this corresponds not providing any theoretical bound (indeed, since all the payoffs are between 0 and 1, any strategy profile has an additive gap no larger than 1). However, even when $\epsilon = 1$, the algorithm terminates only for $m\leq 50$ with $n=3$, for $m\leq 15$ with $n=4$, and for $m=5$ with $n=5$, that is a strictly smaller subset of instances than the subset solved by BARON. The quality of the solutions returned by the algorithm is rather good, but it is always worse than the solutions returned by BARON. Finally, notice that, differently from the previous algorithm, in the case of $n=3$, the quality of the solution decreases as the number of action increases. 

Let us focus on the iterated linear programming. We report the performance of the algorithm for a number of random restarts in $\{1,10,20,30,40,50,60,70,80,90,100\}$. This algorithm solves all the instances of our experimental settings, including also the instances that BARON does not solve due to memory limits. Also in this case, the quality of the solutions is always worse than the solutions returned by BARON. Notice that the ratio is very high and very close to 1 for $n=3$. Interestingly, the number of restarts affects the solution quality essentially only when $m$ is small. Obviously, when $m$ is large, the number of restarts performed by the algorithm reduces, but, surprisingly, the solution quality increases and it is almost the same for every number of restarts. We observe that the number of restarts performed with the largest instances is 30 with $n \in \{3,4\}$ and 10 with $n=5$. This algorithm provides the best approximation w.r.t. the previous two algorithms.

\subsubsection{Summary} Global optimization (BARON) provides the best approximate solutions, but it does not solve all the instances of the experimental setting due to memory limits. The algorithm iterating linear programming allows one to solve larger instances with a relatively small loss in terms of utility. Furthermore, approximating the Team--maxmin equilibrium gets easier as the number of actions per player increases. We observe that this may happen because, as the number of actions per player increases, the probability that the Team--maxmin equilibrium has a small support increases and small supported equilibria should be easier to be found.

\subsubsection{Price of Uncorrelation}
\begin{figure}[h]
\begin{scriptsize}
\centering
  \subfigure[Average] {\includegraphics[width=0.5\textwidth]{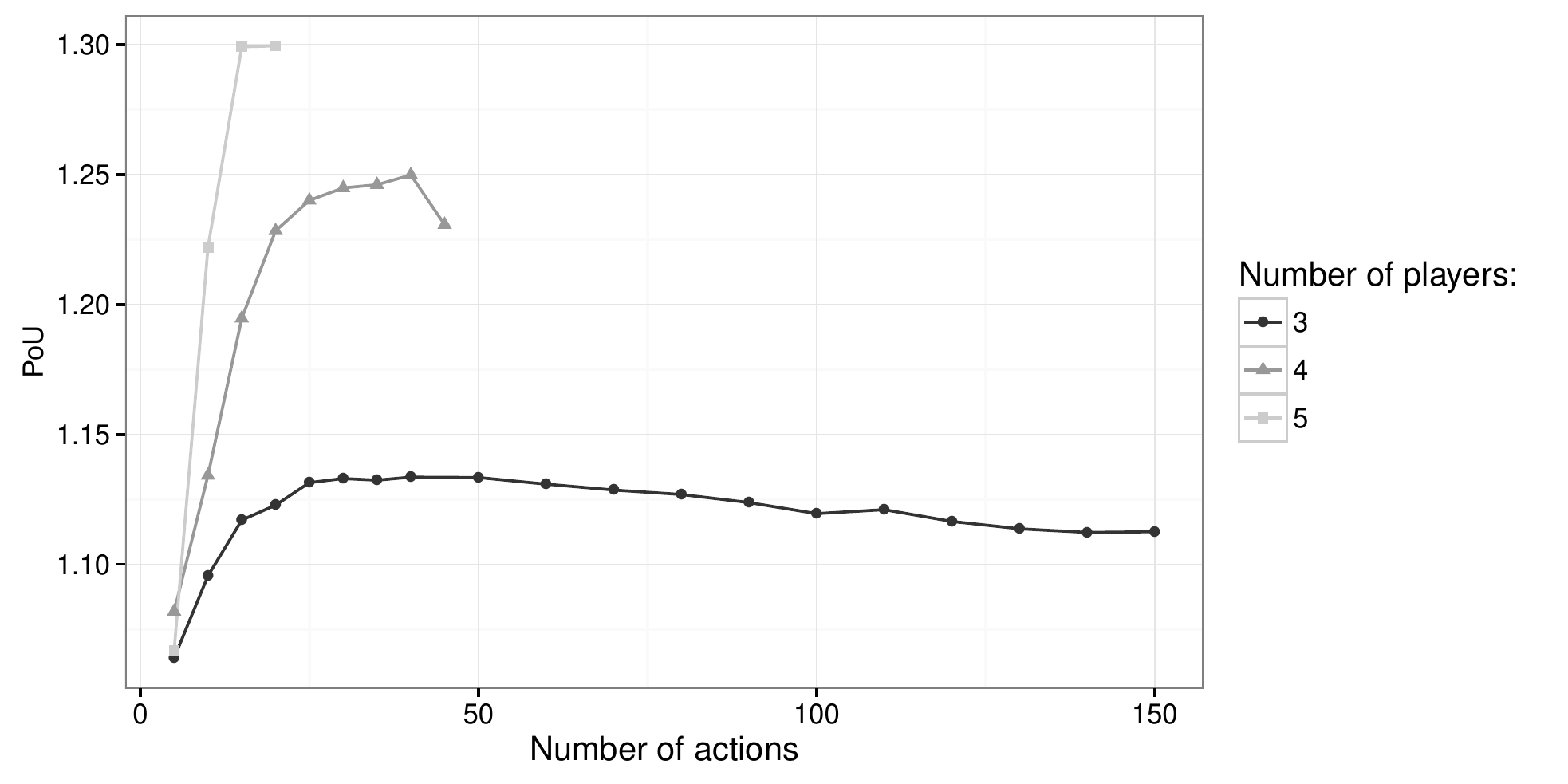}\label{fig:pou_avg}}
  \subfigure[Boxplot] {\includegraphics[width=0.5\textwidth]{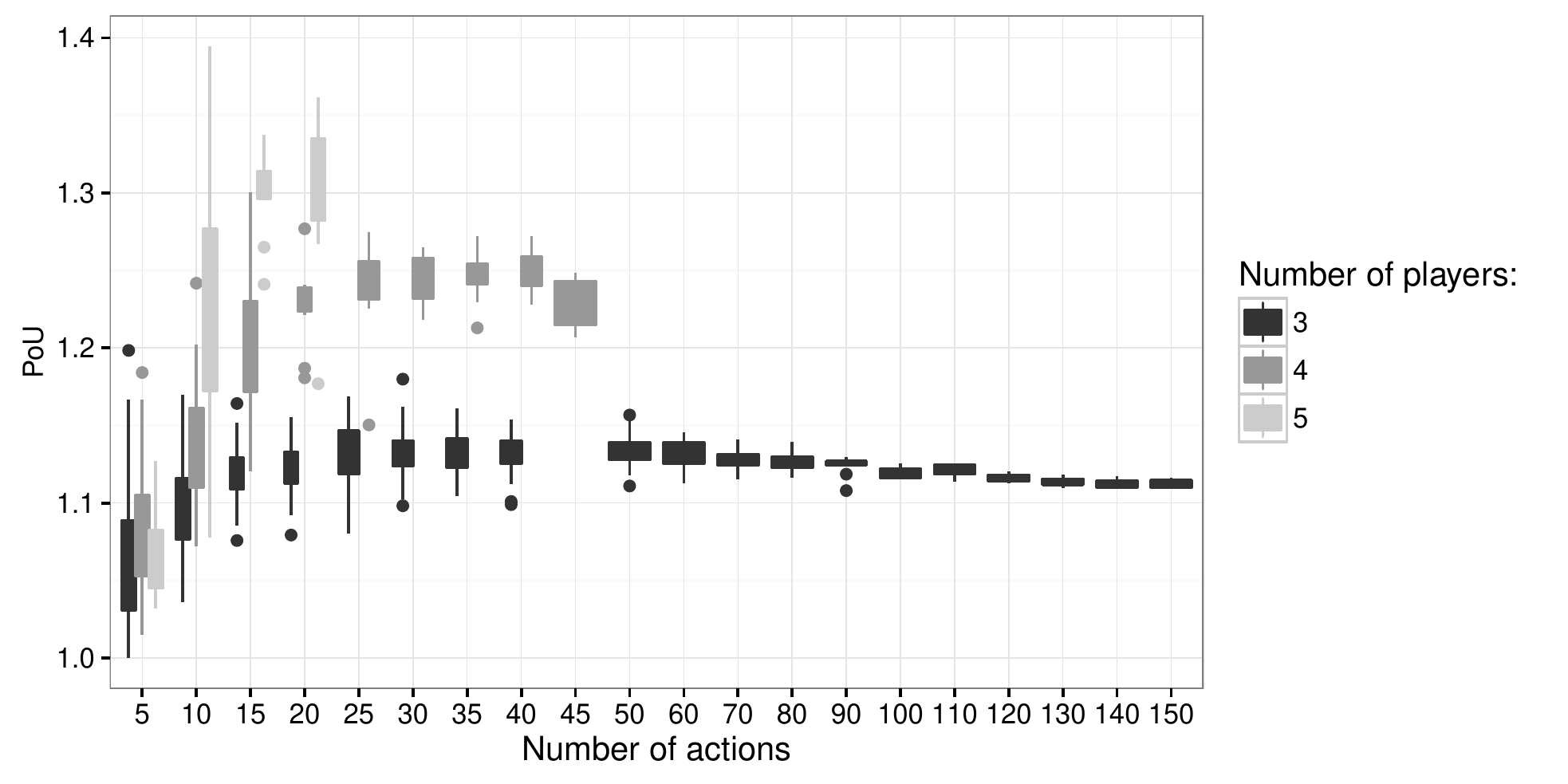}\label{fig:pou_box}}
\caption{Empiric Price of Uncorrelation.}\label{fig:non_corr}
\end{scriptsize}
\end{figure}

Finally, we empirically evaluate \textsc{PoU} in our experimental setting. We do that by calculating the ratio between the value of the Correlated--team maxmin equilibrium---computed exactly---and the lower bound retuned by BARON, which is the algorithm returning always the best approximation of the Team--maxmin equilibrium. This ratio is obviously an upper bound of the actual \textsc{PoU}. In Fig.~\ref{fig:non_corr}, we report the average ratio and the corresponding box plot. Surprisingly, the ratio is very close to 1 even for $n=5$, while, we recall, the worst--case ratio is $m^{n-2}$. It can be observed that the ratio is monotonically increasing in $n$, while the dependency on $m$ is not monotone: there is a maximum small values of $m$ and then the ratio decreases as $m$ increases. For instance, in 3--players games, the ratio goes asymptotically to about 1.15, showing that the loss is very small empirically. Unexpectedly, this shows that, on average, the loss due to the inability for a team of correlating their strategies is rather small.

\subsubsection{Specific game classes} 
Our Appendices includes additional experiments assessing, analogously to what done in this section with \texttt{RandomGames}, empirical approximation performances and Price of Uncorrelation.

These additional results substantially confirm global optimization as the best approach for games of small size, while \texttt{IteratedLP} keeps providing an approximation ratio close to the optimum even with complex game instances. 
Moreover, by generating random games from specific classes, it is possible to observe an interesting set of worst--case instances from the class \texttt{Travelers Dilemma}. 
For those games BARON, although keeping to be the best among the other algorithms, provided low empirical approximation factors suggesting how Team--maxmin could be hard to approximate for such class. 
The Price of Uncorrelation shows a similar trend of that of Fig.~\ref{fig:non_corr} for all classes but \texttt{Bertrand Oligopoly}. For these games, results seem to suggest that correlation becomes more and more critical as the number of players' actions increases.

\section{Conclusions}
\label{sec:conclusions}

The Team--maxmin equilibrium is an important solution concept  requiring deep algorithmic studies. In this work, we studied its efficiency w.r.t. Nash equilibrium and the Maxmin equilibrium with correlated team strategies. Moreover, we proposed algorithms to compute/approximate it, deriving theoretical guarantees and empirical evaluations. 

In future, we will deal with Team--maxmin equilibrium in specific games like polymatrix games and congestion games.

\bibliographystyle{aaai}
\bibliography{biblio}

\clearpage

\appendix
\section*{Appendices}
\section{Proofs of the Theorems}

\begin{theorem} The Price of Anarchy (\textsc{PoA}) of the Nash equilibrium w.r.t. the Team--maxmin equilibrium may be \textsc{PoA}$ =\infty$ even in games with 3 players (2 teammates), 2 actions per player, and binary payoffs for the team.
\label{thm:POA}
\end{theorem}

\noindent 
\emph{Proof}. Consider the following game instance with 3 players (2 teammates), 2 actions per player, and binary payoffs for the team:
\begin{equation*}
\begin{array}{rc|c|c|c|}
\multicolumn{2}{r}{}&\multicolumn{2}{c}{\textnormal{2}}\\
&	& \mathsf{a_3}		& \mathsf{a_4}		\\\cline{2-4}
\multirow{2}{*}{\begin{sideways}\textnormal{1}\end{sideways}}	&\mathsf{a_1} & 1	& 0			\\ \cline{2-4}
&	\mathsf{a_2}	& 0			& 0	\\ \cline{2-4}
\end{array}
\hspace{1cm}
\begin{array}{rc|c|c|c|}
\multicolumn{2}{r}{}&\multicolumn{2}{c}{\textnormal{2}}\\
&	& \mathsf{a_3}		& \mathsf{a_4}		\\\cline{2-4}
\multirow{2}{*}{\begin{sideways}\textnormal{1}\end{sideways}}	&\mathsf{a_1} & 0	& 0			\\ \cline{2-4}
&	\mathsf{a_2}	& 0			& 1	\\ \cline{2-4}
\end{array}
\end{equation*}
\begin{equation*}
\hspace{1.25cm}\mathsf{a_5}\hspace{3.3cm}\mathsf{a_6}
\end{equation*}
\begin{equation*}
\hspace{1cm}3
\end{equation*}

\emph{Team--maxmin equilibrium}. Consider the strategy profile in which:
\[
s_1 = 
\begin{cases}
\mathsf{a}_1	& 0.5	\\
\mathsf{a}_2	& 0.5	\\
\end{cases},
\hspace{0.5cm}
s_2 = 
\begin{cases}
\mathsf{a}_3	& 0.5	\\
\mathsf{a}_4	& 0.5	\\
\end{cases},
\hspace{0.5cm}
s_3 = 
\begin{cases}
\mathsf{a}_5	& 0.5	\\
\mathsf{a}_6	& 0.5	\\
\end{cases}.
\]
It is a Nash equilibrium providing each teammate a utility of 0.25. Indeed, player~$1$ is indifferent between playing actions $\mathsf{a}_1$ and $\mathsf{a}_2$, each action providing a utility of 0.25. The same holds for the other two players: player~$2$ is indifferent between playing actions $\mathsf{a}_3$ and $\mathsf{a}_4$, each action providing a utility of 0.25, and player~$3$ is indifferent between playing actions $\mathsf{a}_5$ and $\mathsf{a}_6$, each action providing a utility of $-0.25$.

\emph{Worst Nash equilibrium}. Consider the action profiles $(\mathsf{a_2}, \mathsf{a_4},\mathsf{a_5})$ and $(\mathsf{a_1}, \mathsf{a_3},\mathsf{a_6})$. They are Nash equilibria providing the team a utility of 0. Indeed, consider $(\mathsf{a_2}, \mathsf{a_4},\mathsf{a_5})$, player~$1$ would gain 0 from unilateral deviations and the same for player~$2$, while player~$3$ would lose utility (from 1 to 0) from unilateral deviations.

\textsc{PoA}. As a result, the ratio between the value of the team provided by the Team--maxmin equilibrium and the worst Nash equilibrium is \textsc{PoA}$=\frac{0.25}{0}=\infty$.\hfill$\Box$

\begin{theorem}
The Price of Uncorrelation (\textsc{PoU}) of the Team--maxmin equilibrium w.r.t. the Correlated--team maxmin equilibrium may be \textsc{PoU}$=m^{n-2}$ even in games with binary payoffs for the team.
\label{thm:POU}
\end{theorem}
\noindent \emph{Proof.} Consider the game instances with $n$ players ($n-1$ teammates) and $m$ actions per player in which the utility of the team is:
\[
U_T(a_1,\ldots,a_n)= 
\begin{cases}
1	&	a_1 = a_2 = \ldots = a_n	\\
0	&	\text{otherwise}
\end{cases},
\]

\emph{Correlated--team maxmin equilibrium}. The equilibrium strategy profile prescribes the team plays all the join actions of the form $a_1=a_2=\ldots=a_{n-1}$, each action played with a probability of $\frac{1}{m}$. The utility for the team is $\frac{1}{m}$. First, we observe that the team cannot improve its utility by playing with strictly positive probability the other joint actions. This easily follows from the fact that all the other actions provide a utility of zero to the team for every action of the adversary, resulting thus weakly dominated. Second, we observe that all the joint actions  of the form $a_1=a_2=\ldots=a_{n-1}$ must be played with strictly positive probability. Indeed, suppose that the team plays with zero probability a joint action of the form $a_1=a_2=\ldots=a_{n-1}=\mathsf{a}$. Then, if the adversary plays $\mathsf{a}$, the team receives a utility of 0. Finally, given that all the joint actions  of the form $a_1=a_2=\ldots=a_{n-1}$ are played with strictly positive probability, the probability over them is uniform by symmetry.

\emph{Team maxmin equilibrium}. The equilibrium strategy profile prescribes each player to play every action with a probability of $\frac{1}{m}$. The utility of the team is $\frac{1}{m^{n-1}}$. Initially, we observe that if some player does not play with strictly positive probability at least an action, say \textsf{a}, then the utility of the team is zero. Indeed, in such a case, if the adversary plays~\textsf{a}, the utility of the team is zero. Finally, given that every player plays with strictly positive probability all the actions, the probability over them is uniform by symmetry.

\textsc{PoU}. As a result, the ratio between the value of the team provided by the Correlated--team maxmin equilibrium and Team--maxmin equilibrium is \textsc{PoU}$=\frac{1/m}{1/m^{n-1}}=m^{n-2}$.\hfill$\Box$

\begin{theorem}
Given any $n$--player game and a Correlated--team maxmin equilibrium with a utility of $v$ for the team, it is always possible to find in polynomial time a mixed strategy profile for the team providing a utility of at least $\frac{v}{m^{n-2}}$ to the team.\label{thm:recostruction}
\end{theorem}
\emph{Proof}. Consider the following simple algorithm. Call $p$ the correlated strategy of the team at the Correlated--team maxmin equilibrium---generically, $p$ is a function $p \in \Delta(A_1 \times \ldots \times A_{n-1})$  where $\Delta(S)$ is the set of all probability distributions over set $S$. The mixed strategy $s_i$ with $i \in N\setminus\{n\}$ is obtained as follows:
\begin{itemize}
\item for every $a_1\in A_1$, we set 
\[
s_1(a_1)=\sum\limits_{(a_2,\ldots a_{n-1})\in A_2 \times \ldots \times A_{n-1}}p(a_1,a_2,\ldots,a_{n-1}).
\]
Notice that the strategy is well defined, summing to 1 since $\sum_{a \in A_1 \times \ldots A_{n-1}}p(a)=1$;
\item for every $i \in N\setminus\{1,n\}$ and for every $a_i \in A_i$, we set 
\[
s_i(a_i)=\begin{cases} 
0 & \not \exists a_{-n} : a_i\in a_{-n} \text{ and } p(a_{-n})>0\\ 
1/|\mathsf{supp}_i| & \text{otherwise}
\end{cases};
\]
where $|\mathsf{supp}_i|$ is the number of actions of~$i$ played with strictly positive probability by $p$. Notice that also in this case the strategy is well defined, summing to 1.
\end{itemize}
In words, the mixed strategy $s_i$ is built such that: $s_1$ plays each action $a_1$ with the probability that $a_1$ is chosen by $p$, while $s_i$ with $i \in N\setminus\{1,n\}$ plays action $a_i$ with a probability of 1 divided by the number of actions of $i$ played by strictly positive probability by $p$ if $a$ is played by $p$ and with a probability of 0 otherwise. We show that the strategy profile $(s_1,\ldots,s_{-n})$ always assures (against an adversary) the team to receive at least $\frac{v}{m^{n-2}}$ where $v$ is the utility given by the Correlated--team maxmin equilibrium. For the sake of the presentation, we distinguish some cases.

\emph{Case 1: for every team member~$i$, each action~$a_i$ is played with positive strictly probability in $p$ and there is only one joint action of the team played with strictly positive probability that contains~$a_i$}. In this case, once an opportune re--labeling of the actions is performed, the correlated strategy $p$ puts strictly positive probability only to joint actions of the form~$a_1 = a_2 = \ldots = a_{n-1}$ and to all such joint actions. First, we show that our $s$ plays each joint action $a_1 = a_2 = \ldots = a_{n-1}$ such that $p(a_1,\ldots,a_{-n})>0$ with $p(a_1,\ldots,a_{-n})/m^{n-2}$. This is because $s_1(a_1) = p(a_1,\ldots,a_{-n})$, while $s_2(a_2)= \ldots = s_{n-1}(a_{n-1})=\frac{1}{m}$. Second, the worst case, which minimizes the utility of the team when a mixed strategy is used, is when the payoffs of all the outcomes achievable with the joint actions played with zero probability by $p$ are zero. This means that the only joint actions providing strictly positive expected utility to the team are those played with strictly positive probability by $p$. Strategy $s$ prescribes over such joint actions the same probability of $p$ divided by $m^{n-2}$. Therefore, the strategy of the adversary does not change and the utility for the team provided by the mixed strategy $s$ is $1/m^{n-2}$ multiplied by the utility given by the Correlated--team maxmin equilibrium. When the payoffs of the outcomes achievable by the joint actions played with zero probability by $p$ have strictly positive values, the expected utility for the team given by $s$ is obviously larger than $1/m^{n-2}$ multiplied by the utility given by the Correlated--team maxmin equilibrium. This completes the proof of the theorem for this case.

\emph{Case 2: for every team member~$i$, each action~$a_i$ is played with positive strictly probability in $p$ and there may be more than one joint action of the team played with strictly positive probability that contains~$a_i$}. In this case, the proof is exactly the same of the previous case except we notice that the probability with which a joint action $(a_1,\dots,a_{-n})$ with $p(a_1,\dots,a_{-n})>0$ is played by $s$ is strictly larger than $\frac{p(a_1,\dots,a_{-n})}{m^{n-2}}$. We provide a simple example. Suppose that action $\mathsf{a}_1$ belongs to 2 joint actions played with strictly positive probability by $p$, say $a'$ and $a''$. The, $s_1(\mathsf{a}_1)= p(a')+p(p'')$. Therefore, the probability with joint action $a'$ is played by $s$ is $\frac{p(a')+p(a'')}{m^{n-2}}>\frac{p(a')}{m^{n-2}}$. Therefore, the utility given to the team by $s$ is, except degeneracies, strictly larger than $\frac{1}{m^{n-2}}$ multiplied by the utility of the Correlated--team maxmin equilibrium.

{\emph{Case 3: no restriction}. In this case, the proof is exactly the same of the previous case except that the probability with which a joint action $(a_1,\dots,a_{-n})$ with $p(a_1,\dots,a_{-n})>0$ is played by $s$ is strictly larger than $\frac{p(a_1,\dots,a_{-n})}{m^{n-2}}$. This follows from the fact that the strategy of team members~$i$ with $i > 1$ may prescribe to play actions~$a_i$ with a probability larger than $1/m$, since $|\mathsf{supp}_i|$ may be smaller than $m$. This completes the proof.
\hfill$\Box$

\begin{theorem}
The \textsc{PoU} of the Team--maxmin equilibrium w.r.t. the Correlated--team maxmin equilibrium may be \textsc{PoU}$ =1$ even in games with binary payoffs.
\label{thm:POUminimo}
\end{theorem}
\noindent \emph{Proof}. To find a suitable example, it suffices to consider a game instance with an equilibrium in pure strategies. Consider, for example, the following game instance with 3 players (2 teammates), 2 actions per player, and binary payoffs for the team:
\begin{equation*}
\begin{array}{rc|c|c|c|}
\multicolumn{2}{r}{}&\multicolumn{2}{c}{\textnormal{2}}\\
&	& \mathsf{a_3}		& \mathsf{a_4}		\\\cline{2-4}
\multirow{2}{*}{\begin{sideways}\textnormal{1}\end{sideways}}	&\mathsf{a_1} & 1	& 0			\\ \cline{2-4}
&	\mathsf{a_2}	& 0			& 1	\\ \cline{2-4}
\end{array}
\hspace{1cm}
\begin{array}{rc|c|c|c|}
\multicolumn{2}{r}{}&\multicolumn{2}{c}{\textnormal{2}}\\
&	& \mathsf{a_3}		& \mathsf{a_4}		\\\cline{2-4}
\multirow{2}{*}{\begin{sideways}\textnormal{1}\end{sideways}}	&\mathsf{a_1} & 1	& 0			\\ \cline{2-4}
&	\mathsf{a_2}	& 0			& 1	\\ \cline{2-4}
\end{array}
\end{equation*}
\begin{equation*}
\hspace{1.25cm}\mathsf{a_5}\hspace{3.3cm}\mathsf{a_6}
\end{equation*}
\begin{equation*}
\hspace{1cm}3
\end{equation*}
Consider any strategy profile where players 1 and 2 play with probability 1 actions $a_1$ and $a_3$, respectively. For any strategy of player 3 this is an equilibrium profile that yields to the team a probability of 1 independently from the availability of correlation between them. \hfill $\Box$

\begin{theorem}
When the algorithm is initialized with a uniform strategy for every player, the worst--case approximation factor of Algorithm~\ref{alg:iterated-lp} is at least $\frac{1}{m^{n-1}}$ and at most $\frac{1}{m^{n-2}}$. When instead the algorithm is initialized with a pure strategy, the worst--case approximation factor is 0.
\end{theorem}
\noindent \emph{Proof}. We prove that, when every team player plays a uniform strategy, the value of the team is almost $\frac{1}{m^{n-1}}$ of the value of the Team--maxmin equilibrium. Consider initially the worst case in which the Team--maxmin equilibrium is pure giving a value $v$ to the team. When a uniform strategy is used by every player, each outcome is played with a probability of at least $\frac{1}{m^{n-1}}$, including the equilibrium outcome. In the worst case, all the outcomes except the equilibrium provide a utility of zero to the team, and therefore the team receives a utility of $\frac{v}{m^{n-1}}$. In the case in which the Team--maxmin equilibrium is mixed, the proof is similar.

Now, we prove that there is some case in which the algorithm returns an approximation of $\frac{1}{m^{n-2}}$. Consider a team game where the team is composed by $n-1$ members and each player has $m$ actions, while the adversary has only one action. The team utility is defined as $U_T(a_1,a_2,\ldots,a_{n-1})=1$ when $a_1=a_2=\ldots=a_{n-1}$ and 0 otherwise. Let us consider, in Line~\ref{alg2:init} of Algorithm~\ref{alg:iterated-lp}, a uniform $\hat{s}$ prescribing to each team member to play each action with probability $\frac{1}{m}$. The resolution of the LP of Line~\ref{alg2:lp}, once fixed the strategy of one team member to such uniform distribution, would output the same uniform distribution for the other one. In other words, $\hat{s}$ is a local maximum for our algorithm. Thus, the algorithm would return a strategy profile guaranteeing the team a payoff of $\frac{m}{m^{n-1}}=\frac{1}{m^{n-2}}$ whereas the team--maxmin value amounts to $1$ and can be clearly obtained by any team profile in which both team members play the same action in pure strategies. So the approximation factor achieved by the algorithm would be exactly $\frac{1}{m^{n-2}}$.

Consider the case in which the initialization is pure. Consider the game instance used in the proof of Theorem~1. If the initialization is $\hat{s}_1 = \mathsf{a}_2$ and $\hat{s}_2 = \mathsf{a}_4$, the algorithm returns 0, while the optimal value is 0.25.\hfill$\Box$

\section{Team--maxmin value irrationality}

In~\cite{DBLP:conf/wine/HansenHMS08}, the authors consider the following 3--player game (we report only the utility of player~3):
\begin{center}
\begin{equation*}
\begin{array}{rc|c|c|c|}
\multicolumn{2}{r}{}&\multicolumn{2}{c}{\textnormal{2}}\\
&	& \mathsf{a_3}		& \mathsf{a_4}		\\\cline{2-4}
\multirow{2}{*}{\begin{sideways}\textnormal{1}\end{sideways}}	&\mathsf{a_1} & 1	& 0			\\ \cline{2-4}
&	\mathsf{a_2}	& 0			& 0	\\ \cline{2-4}
\multicolumn{2}{r}{}&\multicolumn{2}{c}{\mathsf{a_5}}\\
\end{array}
\hspace{2cm}
\begin{array}{rc|c|c|c|}
\multicolumn{2}{r}{}&\multicolumn{2}{c}{\textnormal{2}}\\
&	& \mathsf{a_3}		& \mathsf{a_4}		\\\cline{2-4}
\multirow{2}{*}{\begin{sideways}\textnormal{1}\end{sideways}}	&\mathsf{a_1} & 0	& 0			\\ \cline{2-4}
&	\mathsf{a_2}	& 0			& 2	\\ \cline{2-4}
\multicolumn{2}{r}{}&\multicolumn{2}{c}{\mathsf{a_6}}\\
\end{array}
\end{equation*}
$3$
\end{center}
and look for the minmax strategy of players~1 and~2 against player~3. The authors claim that such a strategy is to play actions~$\mathsf{a_1}$ and~$\mathsf{a_3}$ with probability~$2-\sqrt{2}$, giving player~3 with a utility of $6-4\sqrt{2}$. This is not true, since, if players~1 and~2 play actions $(\mathsf{a}_1,\mathsf{a}_4)$ or $(\mathsf{a}_2,\mathsf{a}_3)$, player~3 gains 0. Therefore, the minmax strategy of players~1 and~2 against player~3 is to play $(\mathsf{a}_1,\mathsf{a}_4)$ or $(\mathsf{a}_2,\mathsf{a}_3)$.

In order to fix the proof, it is sufficient to change the sign of the payoffs. Once the sign has been changed, the example works for both the minmax strategy of players~1 and~2 against player~3 and the team--maxmin equilibrium where players~1 and~2 compose the team. We report here all the calculations---omitted in~\cite{DBLP:conf/wine/HansenHMS08}---for the case of the team--maxmin equilibrium. To compute the team--maxmin strategy of the team we need to solve the following optimization problem:
\begin{equation*}
\begin{array}{ll}
\max\limits_{s_1(\mathsf{a}_1),s_2(\mathsf{a}_3)}& v\\
\text{s.t.} & v  \leq s_1(\mathsf{a}_1)s_2(\mathsf{a}_3)\\
& v  \leq 2(1-s_1(\mathsf{a}_1))(1-s_2(\mathsf{a}_3))\\
& s_1(\mathsf{a}_1) \in [0,1]	\\
& s_2(\mathsf{a}_3) \in [0,1]	
\end{array}
\end{equation*}
Here, the maximum value is achieved when both inequalities hold as equalities. Thus, we can write:
\[
s_1(\mathsf{a}_1)s_2(\mathsf{a}_3) = 2(1-s_1(\mathsf{a}_1))(1-s_2(\mathsf{a}_3))
\]
from which it follows:
\[
s_1(\mathsf{a}_1) = \frac{2s_2(\mathsf{a}_3)-2}{s_2(\mathsf{a}_3)-2}.
\]
Now we write:
\begin{equation*}
v = \frac{s_2(\mathsf{a}_3)(2s_2(\mathsf{a}_3)-2)}{s_2(\mathsf{a}_3)-2}.
\end{equation*}
This expression is maximized for $s_2(\mathsf{a}_3) = 2\sqrt{2}$ and the corresponding value is $6-4\sqrt{2}$.

\newpage
\section{Additional experimental results}

In this section we report additional experimental results we obtained with specific classes of GAMUT games~\cite{nudelman2004run}. We considered 3--Players games from the following classes: \texttt{Bertrand Oligopoly}, \texttt{Dispersion Game}, \texttt{Minimum Effort Game}, and \texttt{TravelersDilemma}. In the following we report the average approximation factor, over 20 random instances for each data point, obtained with BARON and the ratio between BARON's lower bound (corresponding to the best feasible solution returned) and the utility value returned by other algorithms. For Iterating LP we also display what obtained with different numbers of restarts.

\clearpage

\begin{figure*}[!htbp]
\begin{scriptsize}
\centering
 \includegraphics[width=0.6\textwidth]{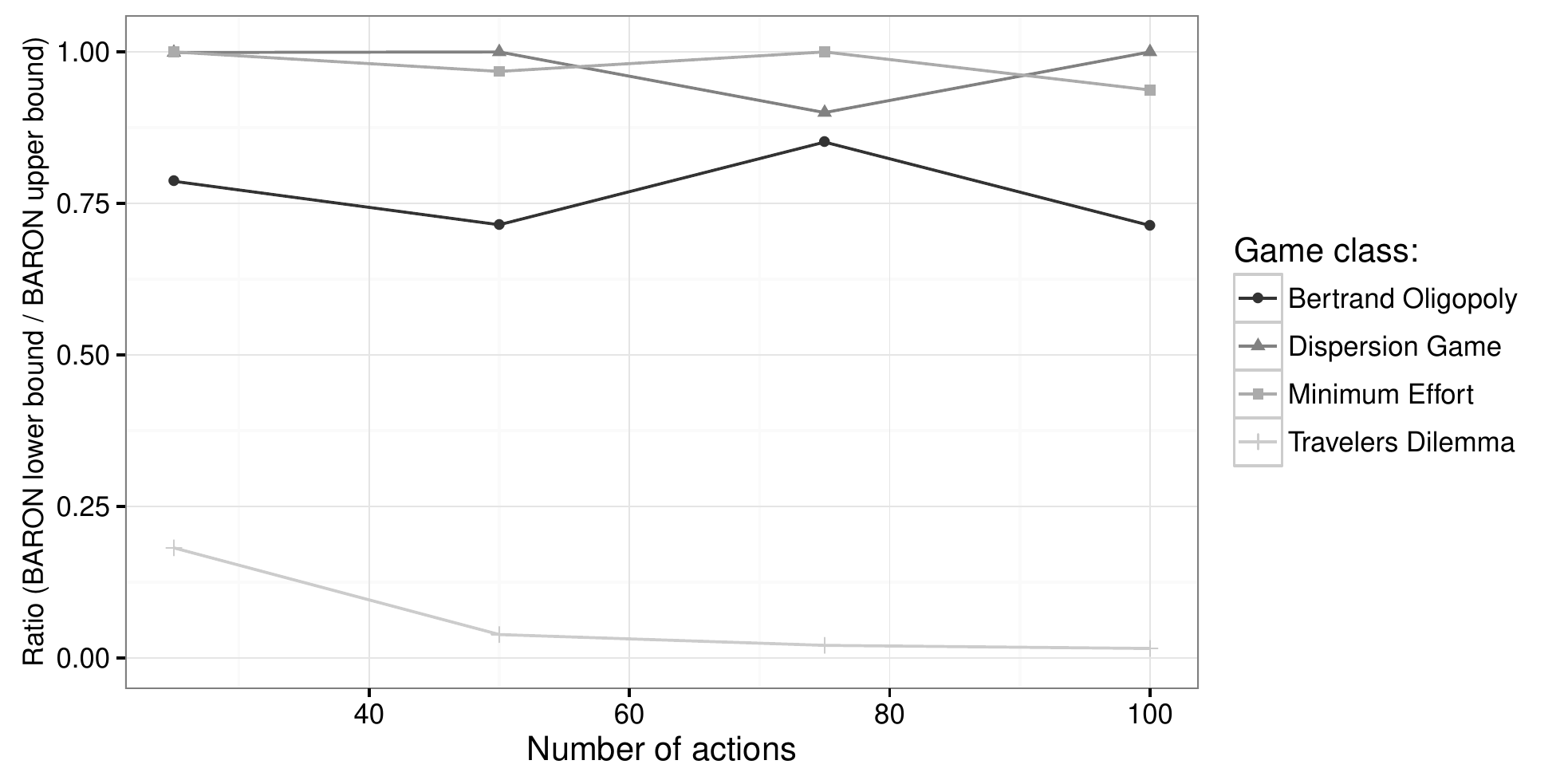}
\caption{Analysis of the approximation ratio obtained with BARON.}
\label{figsup:baron}
\end{scriptsize}
\end{figure*}

\begin{figure*}[!htbp]
\begin{scriptsize}
\centering
 \includegraphics[width=0.6\textwidth]{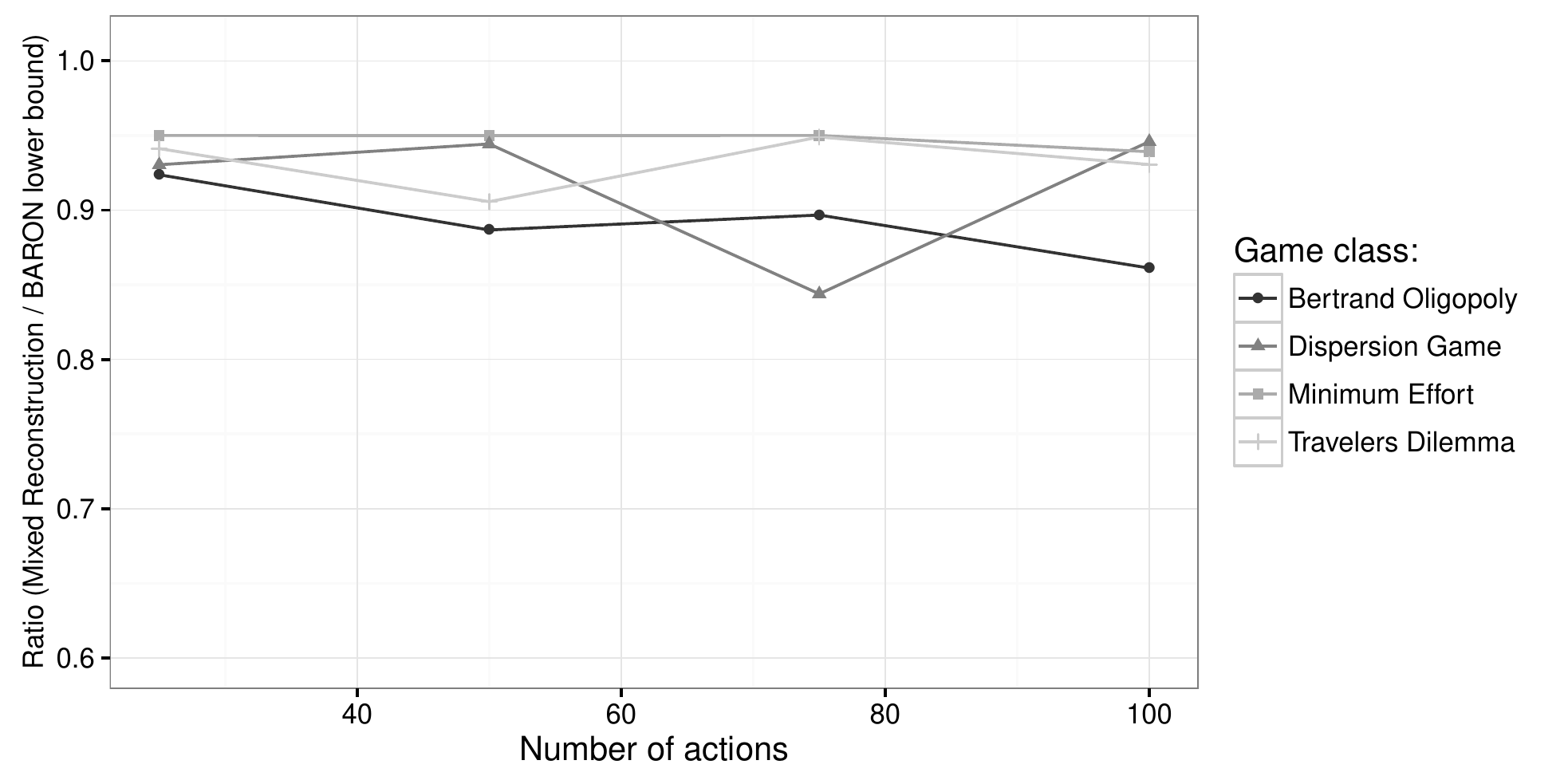}
\caption{Average approximation performance for Reconstrunction from correlated strategies w.r.t. BARON.}
\label{figsup:recon}
\end{scriptsize}
\end{figure*}

\begin{figure*}[!htbp]
\begin{scriptsize}
\centering
 \includegraphics[width=0.6\textwidth]{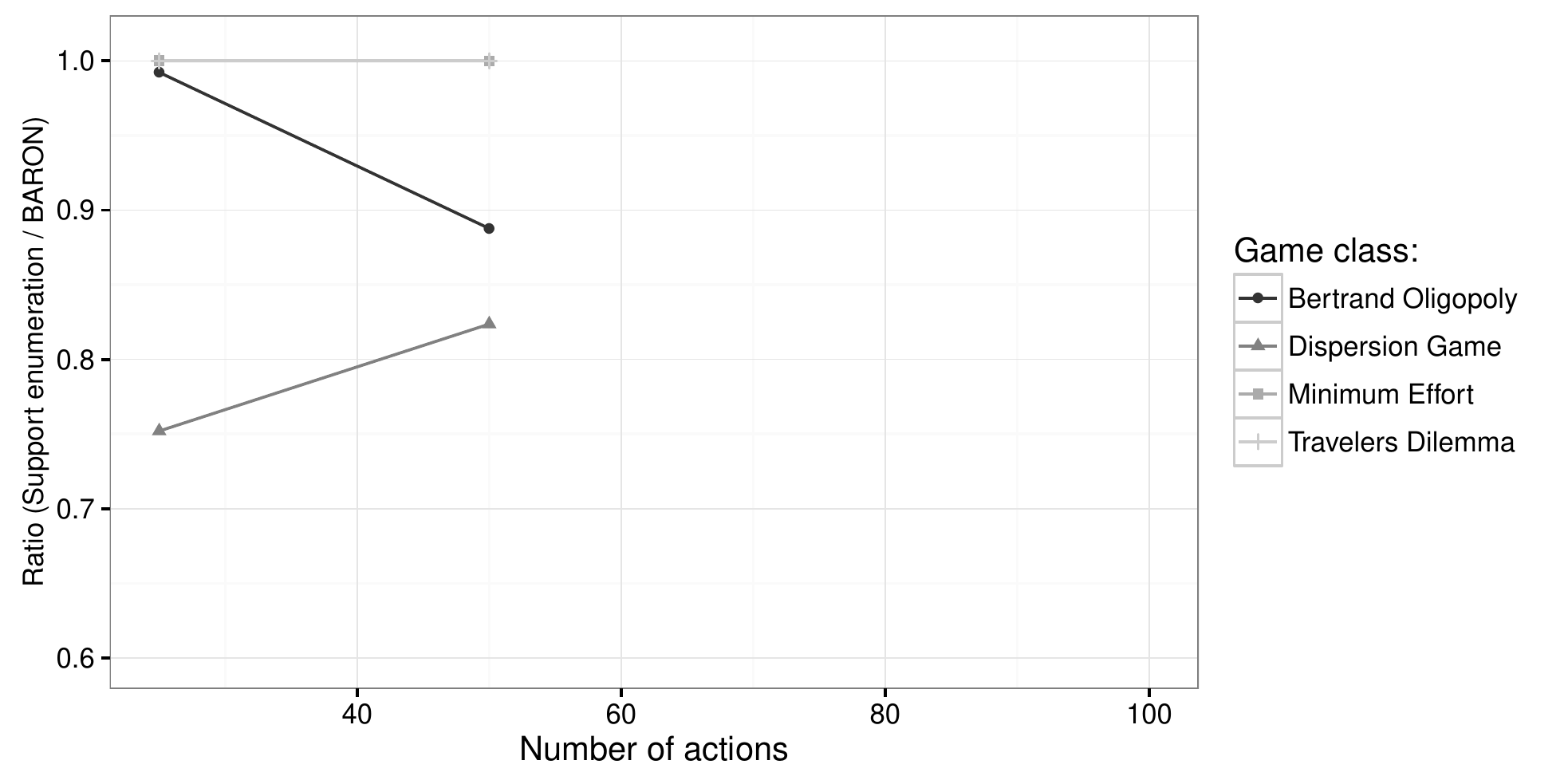}
\caption{Average approximation performance for Support enumeration w.r.t. BARON.}
\label{figsup:recon}
\end{scriptsize}
\end{figure*}

\clearpage

\begin{figure*}[!tbp]
\begin{scriptsize}
\centering
\subfigure[Bertrand Oligopoly.]
 {\includegraphics[width=0.48\textwidth]{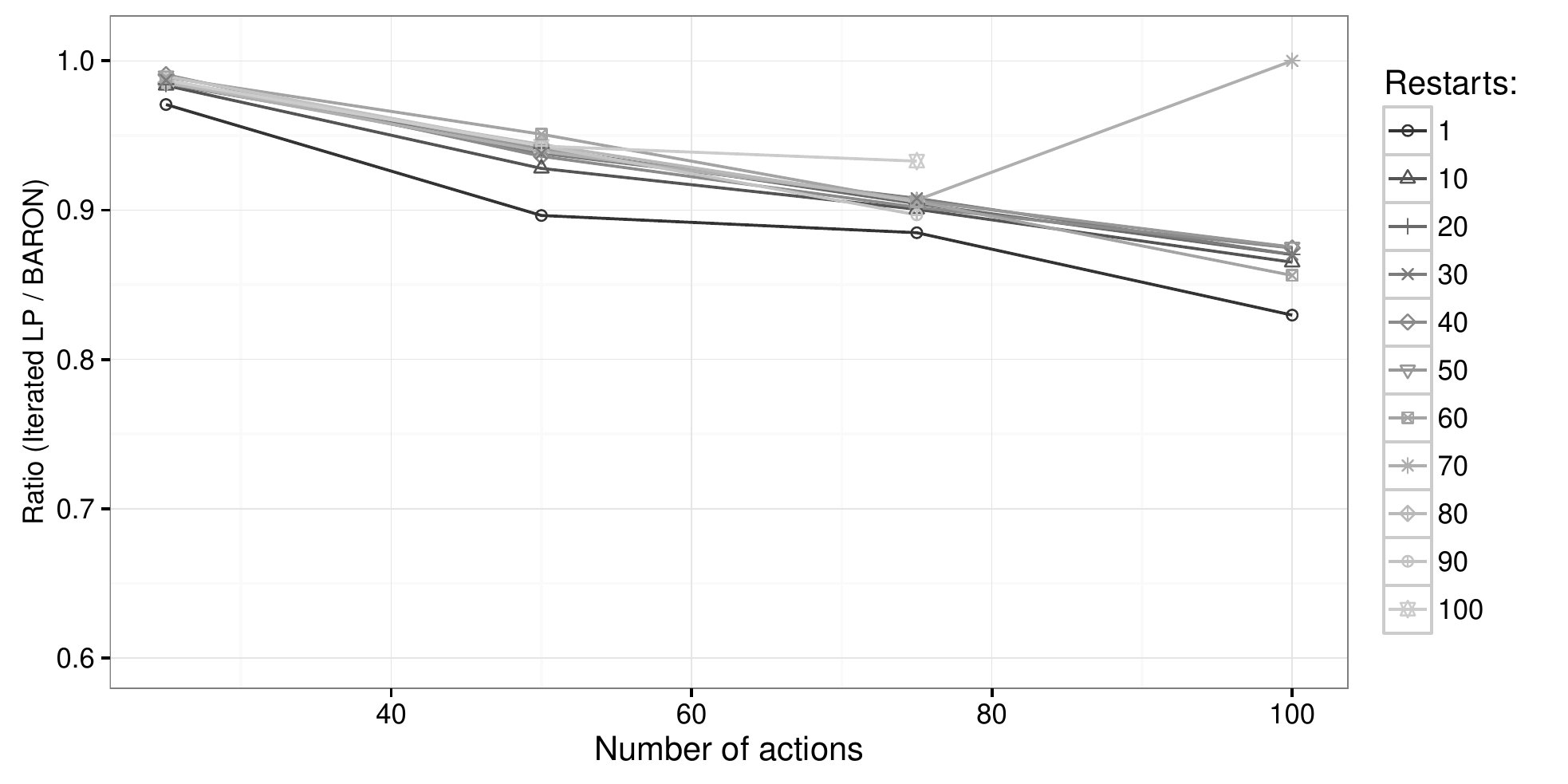}}
\subfigure[Dispersion Game.]
 {\includegraphics[width=0.48\textwidth]{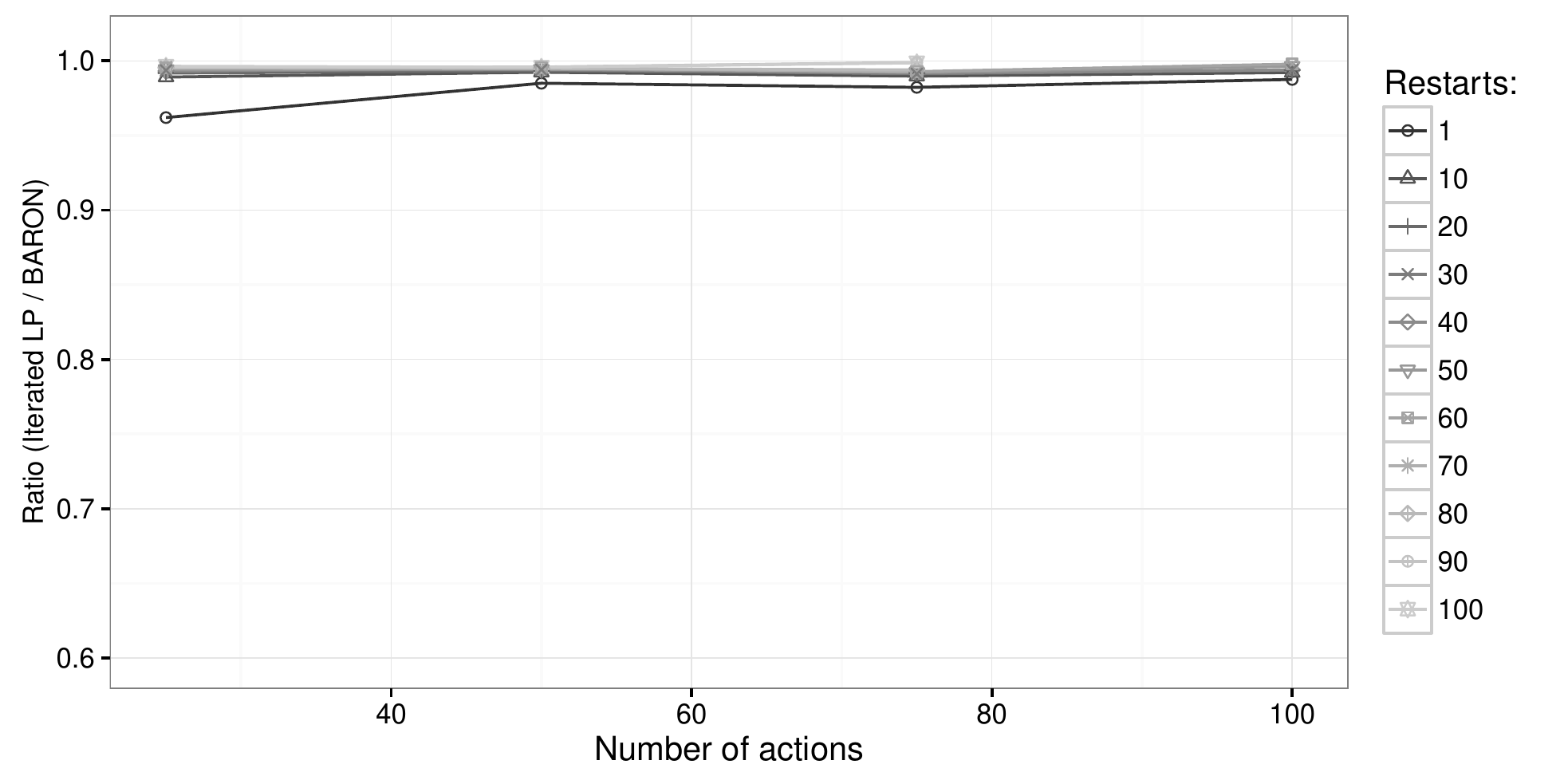}}
\subfigure[Minimum Effort Game.]
 {\includegraphics[width=0.48\textwidth]{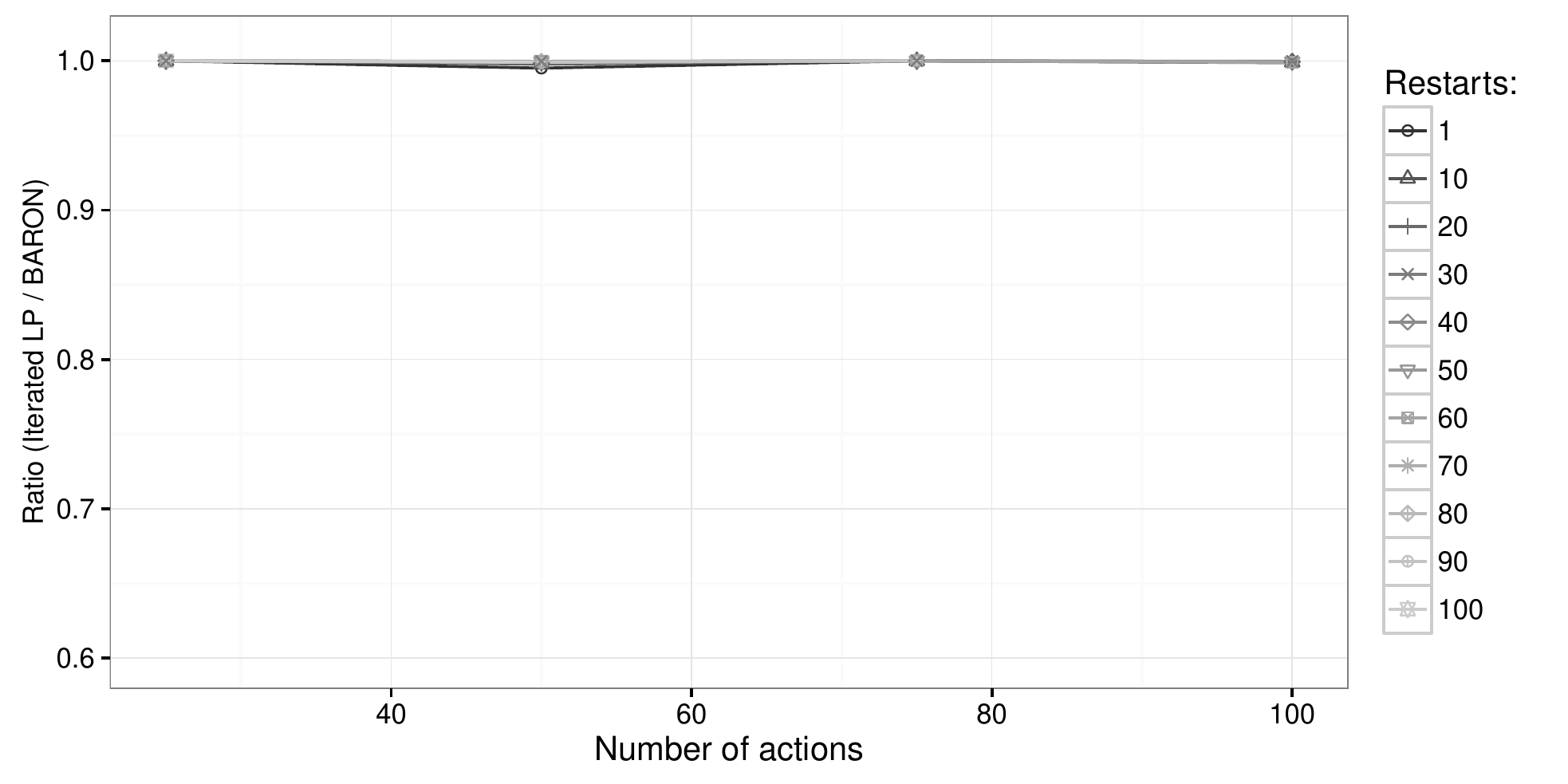}}
\subfigure[TravelersDilemma.]
 {\includegraphics[width=0.48\textwidth]{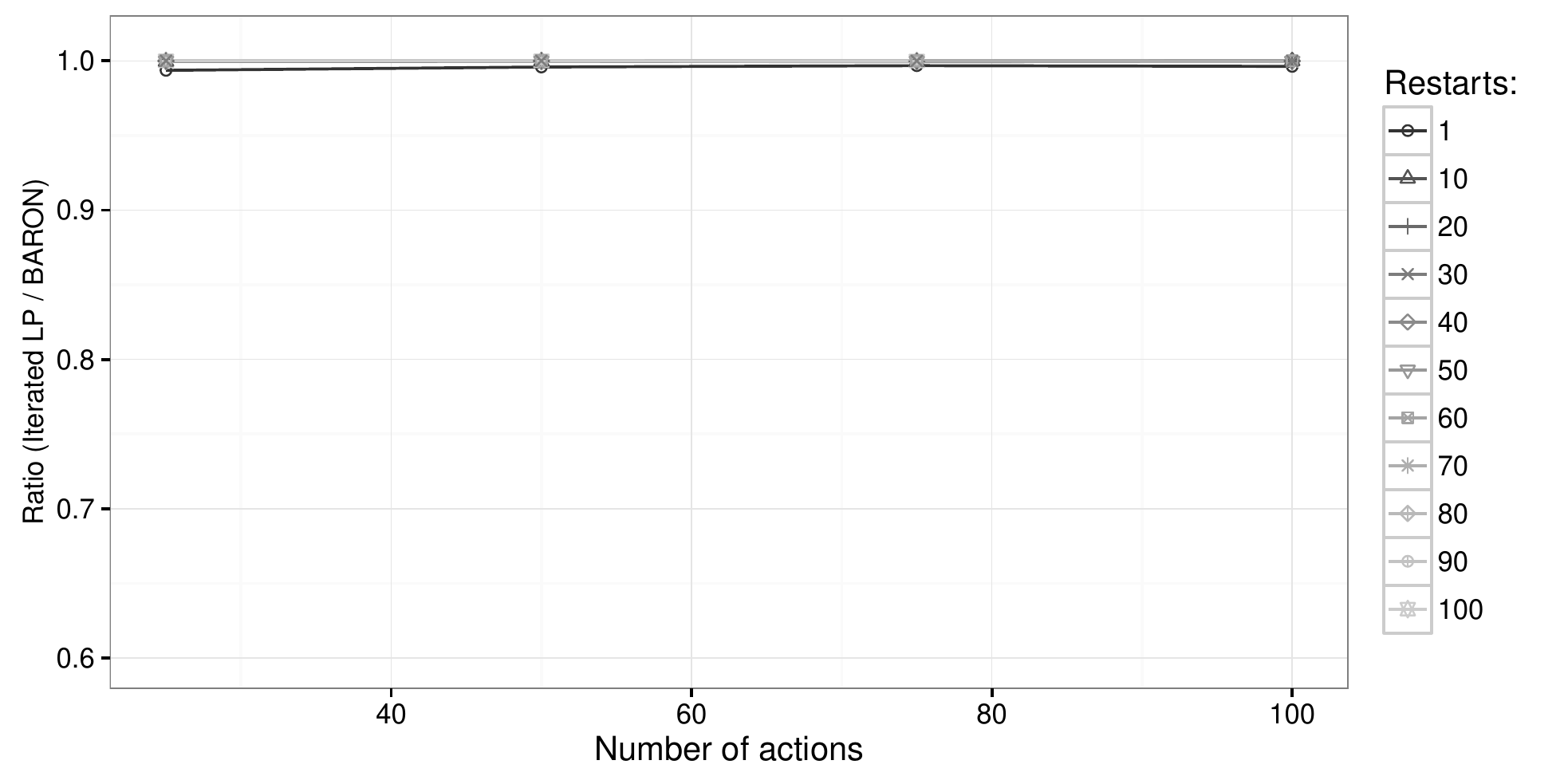}}
\caption{Average approximation performance for Iterating linear programming w.r.t. BARON.}
\label{figsup:iteratingLP}
\end{scriptsize}
\end{figure*}

\begin{figure*}[!htbp]
\begin{scriptsize}
\centering
 \includegraphics[width=0.6\textwidth]{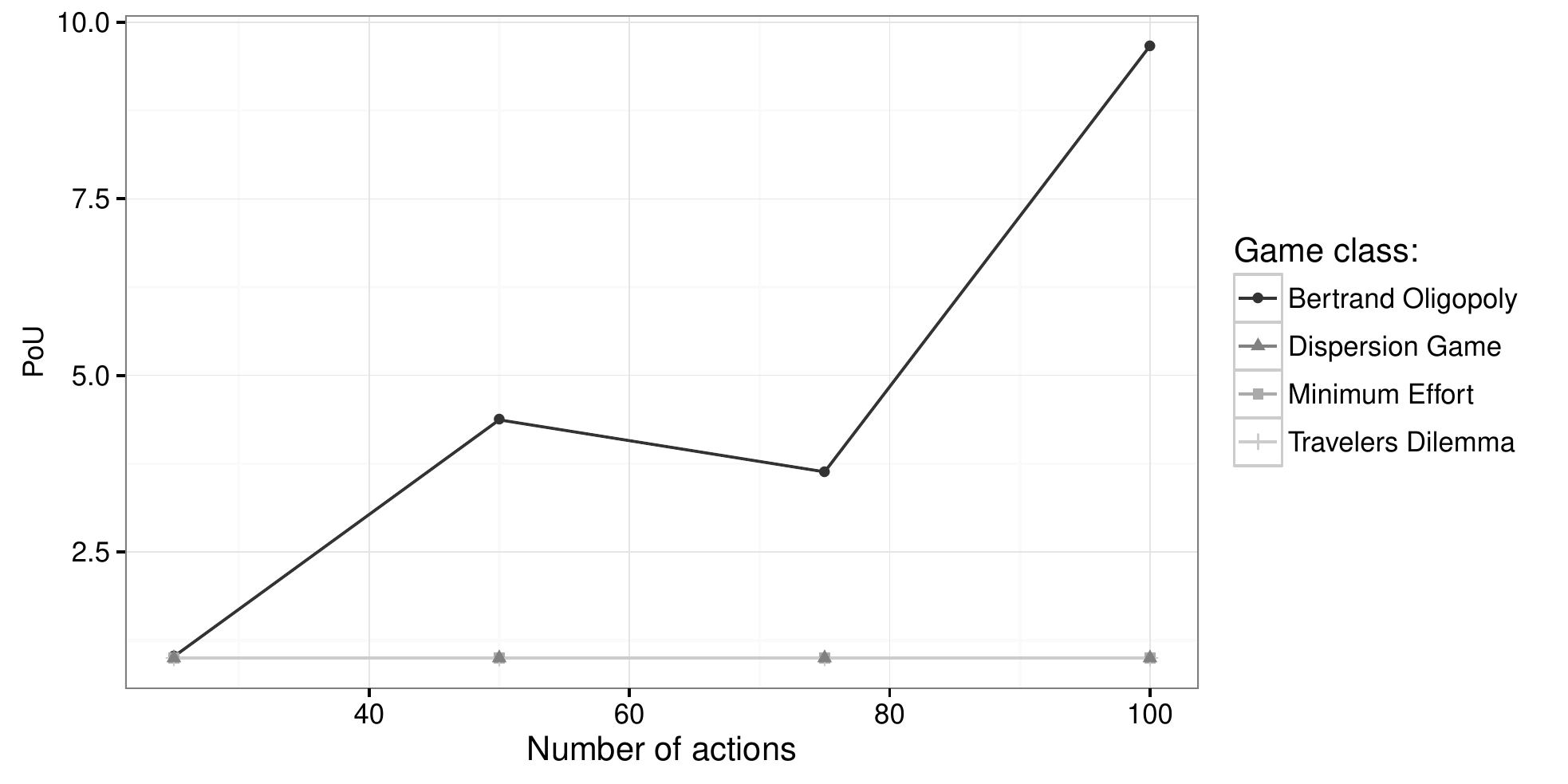}
\caption{Empiric Price of Uncorrelation.}
\label{figsup:pou}
\end{scriptsize}
\end{figure*}

\end{document}